\documentclass[conference]{IEEEtran}
\ifCLASSINFOpdf
\else
\fi
\usepackage{amssymb}
\usepackage{amsmath}
\usepackage{url}
\usepackage[pdftex]{graphicx}
\usepackage{array}
\usepackage{caption}
\usepackage{subcaption}
\usepackage{slashbox}
\usepackage{tikz}
\newcolumntype{?}{!{\vrule width 2pt}}

\begin{document}
%
\title{An Efficient Large-scale Semi-supervised Multi-label Classifier Capable of Handling Missing labels}

\author{
\IEEEauthorblockN{Amirhossein Akbarnejad}
\IEEEauthorblockA{Department of Computer Engineering
\\Sharif University of Technology\\
Tehran , Iran\\
akbarnejad@ce.sharif.edu}
\and
\IEEEauthorblockN{Mahdieh Soleymani Baghshah}
\IEEEauthorblockA{Department of Computer Engineering
\\Sharif University of Technology\\
Tehran , Iran\\
soleymani@sharif.edu}
}


%


\maketitle

\begin{abstract}
Multi-label classification has received considerable interest in recent years. Multi-label classifiers have to address many problems including: handling large-scale datasets with many instances and a large set of labels, compensating missing label assignments in the training set, considering correlations between labels, as well as exploiting unlabeled data to improve prediction performance. To tackle datasets with a large set of labels, embedding-based methods have been proposed which seek to represent the label assignments in a low-dimensional space. Many state-of-the-art embedding-based methods use a linear dimensionality reduction to represent the label assignments in a low-dimensional space. However, by doing so, these methods actually neglect the tail labels - labels that are infrequently assigned to instances. We propose an embedding-based method that non-linearly embeds the label vectors using an stochastic approach, thereby predicting the tail labels more accurately. Moreover, the proposed method have excellent mechanisms for handling missing labels, dealing with large-scale datasets, as well as exploiting unlabeled data. With the best of our knowledge, our proposed method is the first multi-label classifier that simultaneously addresses all of the mentioned challenges. Experiments on real-world datasets show that our method outperforms state-of-the-art multi-label classifiers by a large margin, in terms of prediction performance, as well as training time.
 
\end{abstract}


%
\IEEEpeerreviewmaketitle

\section{Introduction}
Unlike the traditional single-label classification in which one instance can have only one label, in multi-label classification tasks, an instance can be associated with a set of labels. Multi-label classifiers have been applied, for instance, in multi-label text
classification \cite{example_text}, automated image annotation \cite{mpu}, protein function prediction \cite{example_protein}, and recognition of facial action units in facial images \cite{faur}. A simple approach is to build an independent binary classification model for each label. This approach is referred to as Binary Relevance (BR) \cite{survey_1}. Although BR \cite{survey_1} has a straightforward approach, it does not consider correlations between labels while they are highly correlated. It has been shown \cite{consistent_mlc} that for some performance metrics, the Bayes optimal multi-label classifier does not consider correlations between labels. However, having finite training samples, many state-of-the-art methods take advantage of correlations between labels and can outperform BR \cite{survey_1}.

Real-word multi-label datasets \cite{mulan,xml_repository} usually have a large set of labels and thus building an independent binary classification model for each label is infeasible. Therefore, BR \cite{survey_1} is not capable of handling a large set of labels. To tackle this problem, embedding-based methods have been proposed. Having $K$ labels, we represent the labels that are associated with an instance
by a binary vector of length $K$ - the \textit{label vector} - where each dimension of
this vector is a binary variable that shows whether a specific label is associated with the instance or not. Embedding-based methods seek to represent the label vector of each instance in a low-dimensional space, the \textit{latent space}, and then use the feature vectors along with independent regression models to predict the representation of an instance in the latent space. Many recent methods like FaIE\cite{faie}, LEML \cite{leml}, LCML \cite{lcml}, PLST \cite{plst} and CPLST \cite{cplst} consider a linear relationship between the label vectors and representations in the lower-dimensional latent space. Note that this approach is equivalent to considering a low-rank linear mapping to transform the input feature vectors to the label vectors. However, by doing so, these methods actually neglect the tail labels which are infrequently assigned to instances. Many real-world multi-label datasets \cite{xml_repository} have thousands of tail labels. Neglecting these tail labels can dramatically affect the prediction performance. To tackle this problem, our proposed method considers a non-linear relationship between the label vectors and the representations in the latent space via an stochastic transformation.

Many real-world multi-label datasets \cite{xml_repository} contain millions of instances, each having millions of features, as well as label sets containing millions of labels.
Dealing with these large-scale datasets is challenging. In fact, dealing with large-scale datasets is beyond the pale of many state-of-the-art methods like SLRM \cite{slrm}, FaIE \cite{faie}, and FastTag \cite{fasttag}. In this paper, we propose a probabilistic model that can handle large-scale datasets. In the proposed method, we model the two mappings mentioned above - one of them transforms the feature space to the latent space and the other one transforms the latent space to the label space - by stochastic transformations drawn from sparse Gaussian processes \cite{spgp}. These Gaussian processes are parameterized by some pseudo-instances. The $\mathcal{O}(N^3)$ computational cost and the $\mathcal{O}(N^2)$ memory requirements of the Gaussian processes (GPs) makes them prohibitive for a large set of instances. However, it is shown that by parameterizing a Gaussian process by pseudo-instances, it is possible to achieve the full GP performance at much lower computational costs \cite{spgp}.  

In real-world multi-label datasets, all valid label assignments are not thoroughly provided by the training set. In fact, in multi-label datasets, labels are often obtained
through crowd sourcing or crawling the web pages. Thus, the
associated labels to an instance can be incomplete.
Indeed, although an image or a text document can be associated with many labels, labelers may provide only a subset of them. This problem is referred to as \textit{missing labels}, \textit{weak labels}, or \textit{partially labeled data}. To address the problem of  missing labels, we used the EEOE framework of our previous work \cite{akpaper1}. Application of the EEOE to our new model confirms that EEOE can be considered as a general framework to handle missing labels. 

In this paper, we propose an embedding-based multi-label classifier that models the transformation that maps the feature space to the latent space, as well as the transformation that maps the latent space to the label space, by two sets of stochastic transformations. In this regard, our method is more flexible than state-of-the-art linear approaches and in terms of prediction performance, outperforms them by a large margin. Furthermore, by modeling these mappings using stochastic transformations, the proposed method addresses the problem of neglecting the tail labels, which is not addressed by many state-of-the-art embedding-based multi-label classifiers. 
One of the most important contributions of this paper is to exploit the idea of parameterizing sparse Gaussian processes by pseudo-instances \cite{spgp} that leads to dramatically decreasing the training time, as well as handling large-scale datasets. Moreover, our method has effective mechanisms to compensate missing labels and to exploit unlabeled instances.

\hfill

\subsection{Notation}
Having $N$ labeled instances in the training set, we denote the representation of the $n$-th labeled instance in the feature space and the latent space by $X^{(n)}$ and $C^{(n)}$ respectively, where $X^{(n)} \in \mathbb{R}^{F}$ and $C^{(n)}\in \mathbb{R}^{L}$. Moreover, we denote the unobserved complete label vector and the observed incomplete label vector of the $n$-th labeled instance by $Z^{(n)}$ and $Y^{(n)}$ respectively, where $Z^{(n)}\in \mathbb{R}^K$ and $Y^{(n)}\in \lbrace 0,1 \rbrace ^{K}$. Note that the vector $Z^{(n)}$ is real valued and its $k$-th dimension shows suitability of the $k$-th label for the $n$-th instance. The role of these random variables will be elaborated in the next sections. Similarly, having $I$ unlabeled instances in the training set, we denote the representation of the $i$-th unlabeled instance in the feature space and the latent space by $X^{(i+N)}$ and $C^{(i+N)}$ respectively, where $X^{(i+N)} \in \mathbb{R}^{F}$ and $C^{(i+N)}\in \mathbb{R}^{L}$. Moreover, we denote the unobserved complete label vector of the $i$-th unlabeled instance by $Z^{(i+N)}$, where $Z^{(i+N)}\in \mathbb{R}^K$. The sigmoid function is also shown as $\sigma\big( \bullet \big)$. Given $\mathbf{A} = \lbrace A_1,...,A_M \rbrace$ as an arbitrary set, $A_{-m}$ shows the set containing all the elements of $\mathbf{A}$ except to $A_m$. Tab. \ref{tab:notations}. summarizes the notations and the symbols which are used in this paper.

\begin{table*}[t]
\centering
\caption{Symbols and notations which are used in this paper.}
\hspace{-0.9cm}
\label{tab:notations}
\begin{tabular}{|p{3.0cm}|p{14.0 cm}|}
\hline
\hline
$N$ & Number of labeled instances in the training set.\\
\hline
$I$ & Number of unlabeled instances in the training set.\\
\hline
$F$ & Number of features.\\
\hline
$K$ & Cardinality of the label set.\\
\hline
$L$ & Dimensionality of the latent space. Usually, the following inequality holds: $L \ll K$.\\
\hline
$X^{(n)} \in \mathbb{R}^F$ & Feature vector of the $n$-th labeled instance in the training set.\\
\hline
$X^{(i+N)} \in \mathbb{R}^F$ & Feature vector of the $i$-th unlabeled instance in the training set.\\
\hline
$Y^{(n)} \in \lbrace 0,1 \rbrace^K$ & Label vector of the $n$-th instance. The $k$-th label is assigned to the $n$-th instance if and only if $y^{(n)}_k=1$\\
\hline
$C^{(n)} \in \mathbb{R}^L$ & Embedded representation of the $n$-th label vector 
$Y^{(n)}$.\\
\hline
$C^{(i+N)} \in \mathbb{R}^L$ & Latent space representation for the unobserved label vector of the $i$-th unlabeled instance.\\
\hline
$Z^{(n)}\in \mathbb{R}^{K}$ & The $k$-th dimension of the vector $Z^{(n)}$ shows suitability of the $k$-th label for the $n$-th labeled instance.\\
\hline
$Z^{(i+N)}\in \mathbb{R}^{K}$ & The $k$-th dimension of the vector $Z^{(i)}$ shows suitability of the $k$-th label for the $i$-th unlabeled instance.\\
\hline
$c^{(n)}_\ell \in \mathbb{R}$ & The $\ell$-th elements of the vector $C^{(n)}$. We denote the elements of the vectors $Y^{(n)}$ and $Z^{(n)}$ accordingly.\\
\hline
\hline
$\mathbf{f_c} = \lbrace f_{c}^{(\ell)} \rbrace_{\ell=1}^{L}$ & $f_{c}^{(\ell)}: \; \mathbb{R}^F \rightarrow \mathbb{R}$ is the stochastic function that maps the vectors $\lbrace X^{(n)}\rbrace_{n=1}^{N+I}$ to the $\ell$-the dimension of the vectors $\lbrace C^{(n)}\rbrace_{n=1}^{N+I}$.\\
\hline
$\mathbf{f_z} = \lbrace f_{z}^{(k)} \rbrace_{k=1}^{K}$ & $f_{z}^{(k)}: \; \mathbb{R}^L \rightarrow \mathbb{R}$ is the stochastic function that maps the vectors $\lbrace C^{(n)}\rbrace_{n=1}^{N+I}$ to the $k$-th dimension of the vectors 
$\lbrace Y^{(n)}\rbrace_{n=1}^{N+I}$.\\
\hline
$M_c$ & Number of pseudo-instances for the stochastic functions in $\mathbf{f_c}$.\\
\hline
$M_z$ & Number of pseudo-instances for the stochastic functions in $\mathbf{f_z}$.\\
\hline
$\mathbf{S_c}= \lbrace s^{(m)}_c \rbrace_{m=1}^{M_c}$ & Set of pseudo-samples from the functions in $\mathbf{f_c}$, where $s^{(m)}_c \in \mathbb{R}^F \;,\; 1\le m \le M_c$.\\
\hline
$\mathbf{S_z} = \lbrace s^{(m)}_z \rbrace_{m=1}^{M_z}$ & Set of pseudo-samples from the functions in $\mathbf{f_z}$, where $s^{(m)}_z \in \mathbb{R}^L \;,\; 1\le m\le M_z$. \\
\hline
$\mathbf{U}_c = \lbrace u_{cm}^{(\ell)}\rbrace_{m=1}^{M_c} {}_{\ell=1}^{L}$ & A set containing values of the functions in $\mathbf{f_c}$ 
at the pseudo-instances. In other words, $\hat{c}^{(n)}_\ell \approx f_c^{(\ell)}(X^{(n)})\;,\;1\le n \le (N+I)$.\\
\hline
$\mathbf{U}_z = \lbrace u_{zm}^{(k)}\rbrace_{m=1}^{M_z} {}_{k=1}^{K}$ & A set containing values of the functions in $\mathbf{f_z}$ 
at the pseudo-instances. In other words, $\hat{z}^{(n)}_k \approx f_z^{(k)}(C^{(n)})\;,\;1\le n \le (N+I)$.\\
\hline
$\kappa(\bullet,\bullet\; ; \sigma_c)$ & A radial-basis function kernel with the smoothing parameter $\sigma_c$.\\
\hline
\end{tabular}
\end{table*}

\subsection{Paper Organization}
The rest of this paper is organized as follows: In Section 2 we review some related works. Section 3 explains the proposed method. In Section 4, we compare the performance of our proposed method to that of some state-of-the-art multi-label classifiers. In the last section, we conclude the paper.

\section{Related Work}
\subsection{Handling a Large Set of Labels}
To deal with a large set of labels, two types of methods have been proposed: label selection methods and label transformation methods. Label selection methods \cite{mlcssp,moplms} assume that the label matrix $\mathbf{Y}_{N\times K}$ can be reconstructed by a small subset of its columns. After selecting $L$ columns of the matrix $\mathbf{Y}_{N\times K}$ and getting the matrix $\mathbf{\tilde{Y}}_{N\times L}$, the prediction tasks which are required will be feasible at acceptable computation costs. Indeed, it can predict the matrix $\mathbf{\tilde{Y}}_{N\times L}$ from the feature matrix $\mathbf{X}_{N\times F}$, and the label matrix $\mathbf{Y}_{N\times K}$ from the matrix $\mathbf{\tilde{Y}}_{N\times L}$) in a resonable time. On the other hand, label transformation methods seek to represent label vectors in an $L$-dimensional space, i.e. the latent space, where $L \ll K$. Recall that the vector $C^{(n)} \in \mathbb{R}^L$ denotes the latent space representation of the label vector $Y^{(n)}$. Some multi-label classifiers like LEML \cite{leml} consider a low-rank linear mapping to transform the feature vector $X^{(n)}$ to the label vector $Y^{(n)}$, which is equivalent to representing the label vectors in a lower dimensional space. Many multi-label classifiers like FaIE \cite{faie}, LEML \cite{leml}, CPLST \cite{cplst} and PLST \cite{plst} consider a linear relationship between the vectors $Y^{(n)}$ and $C^{(n)}$. However, by doing so, the tail labels will be omitted from the training process. Indeed, using a linear dimensionality reduction to map the label vectors to the latent space actually fades the tail labels. In this regard, many methods have been proposed to avoid the problem of fading the tail labels. For instance, REML \cite{reml} considers a full-rank linear mapping to predict the tail labels from the feature vectors, while considers a low-rank linear mapping to predict other labels from feature vectors, or SLEEC \cite{sleec} attempts to preserve the distance of a label vector to only a few of its nearest neighbors, thereby modeling the label vectors by a low-dimensional manifold. To avoid the problem of fading the tail labels, our proposed method models the decoder transformation that maps the latent space representations to the label vectors by a set of stochastic transformations, which are drawn from some sparse Gaussian processes \cite{spgp}.

\subsection{Handling Missing Labels}
In multi-label classification, valid label assignments are not thoroughly provided by the training set. This problem - referred to as the presence of missing labels - is more drastic in large-scale datasets. However, many state-of-the-art methods like REML \cite{reml}, SLEEC \cite{sleec} and FaIE \cite{faie} have no explicit mechanisms to handle missing labels.
The common approaches to tackle this problem are as follows: 
(1) Some methods, like 
LEML \cite{leml} and WELL \cite{well}, seek to generate all the ones in the label matrix $\mathbf{Y}$,
rather than attempting to generate both the ones and the zeros of $\mathbf{Y}$.
(2) LCML \cite{lcml} assumes that
the elements of the label matrix $\mathbf{Y}$ are of three types: zero, one and
unknown. Afterwards, it considers the unknown elements as latent variables.
(3) MPU \cite{mpu} is a probabilistic model whose generative process is as follows: Given the representation of instances in the feature space, MPU firstly generates the unobserved complete label vectors of the instances. Then, it removes some labels from the complete label vectors to generate the observed incomplete label vectors. Although this approach is related to ours, we will explain later that our approach is preferable to that of MPU. 
(4) Matrix completion methods have also been applied to multi-label classification tasks. These methods, like IrMMC \cite{irmmc} and MC-1 \cite{mc1}, firstly make a matrix containing all the feature vectors and the label vectors of both the training and the test data. Afterwards, they exploit matrix completion methods to fill in the missing entries of this matrix. These missing entries include label vectors of the test set, missing labels in the training set, as well as unknown features. Thus, these methods can handle missing labels and unknown features.  
(5) FastTag \cite{fasttag} assumes that an observed incomplete label vector can be linearly transformed to the unobserved complete one. FastTag learns this linear mapping via the idea of training a denoising auto-encoder. More precisely, it removes some of the one entries from the label vectors, then assumes that there exist a linear mapping that can retrieve the original label vectors from the corrupted ones. 
(6) In the context of recommender systems, a problem - referred to as \textit{missing rates}- arises which is similar to the problem of missing labels in multi-label classification. Indeed, in these systems, user ratings are usually unknown or missing since a user may have not seen or rated many items. Many approaches have been proposed that are capable of handling missing rates, two of which are Poisson matrix factorization \cite{pmf} and CTR \cite{ctr}. 
\subsection{Dealing with Large-scale Datasets}
In multi-label classification, dealing with large-scale datasets has received considerable interest in recent years. The common approaches to tackle this problem are as follows:
(1) Some multi-label classifiers like MPU \cite{mpu} use a stochastic optimization, thereby dealing with small subsets of the training data, rather than considering whole of it at once. 
(2) It has been shown that by making certain assumptions, it is possible to divide the large-scale multi-label classification task into simpler multi-label classification subtasks so that minimizing the zero-one loss for these subtasks is equivalent to minimizing the zero-one loss for the main multi-label classification task \cite{opt}. However, in the presence of missing labels, minimizing the zero-one loss seems to be inappropriate.
(3) LEML \cite{leml} uses a low-rank linear mapping to transform the feature vectors to the label vectors. To impose the low-rank constraint on this linear mapping, LEML \cite{leml} assumes that the matrix of this transformation can be factorized into a product of two matrices, then exploits sparsity in the feature matrix $\mathbf{X}$ and learns these two matrices in an alternating minimization schema.      
(4) REML \cite{reml} assumes that the label vector ${Y}^{(n)}$ can be decomposed as ${Y}^{(n)} = \hat{{Y}}^{(n)}_{S} + \hat{{Y}}^{(n)}_{L}$, where $\hat{{Y}}^{(n)}_{S}$ contains the tail label assignments. Afterwards, REML \cite{reml} uses a low-rank linear mapping to transform the feature vector $X^{(n)}$ to the vector $\hat{{Y}}^{(n)}_{L}$, while uses a full-rank one to transform the feature vector $X^{(n)}$ to the vector $\hat{{Y}}^{(n)}_{S}$. By doing so, the tail labels will not be faded. Moreover, REML \cite{reml} uses an alternating minimization schema along with a divide-and-conquer approach, thereby handling large-scale datasets.
(5) SLEEC \cite{sleec} seeks to find the latent representations of the label vectors, the vectors $\lbrace C^{(n)} \rbrace _{n=1}^N$, with two desired properties: 1. Similarity between each label vectors and its few nearest neighbors should be preserved in the latent space. 2. Having the feature vectors $\lbrace X^{(n)}\rbrace _{n=1}^N$, it should be possible to predict the vectors $\lbrace C^{(n)} \rbrace _{n=1}^N$. To handle large-scale datasets, SLEEC \cite{sleec} clusters the training data into some clusters, learns a separate embedding per cluster and performs kNN classification within the test point's cluster alone \cite{sleec}. 
\section{Proposed Method}
\subsection{The Model}
\begin{figure}[t]
\centering
\includegraphics[width=3.5in]{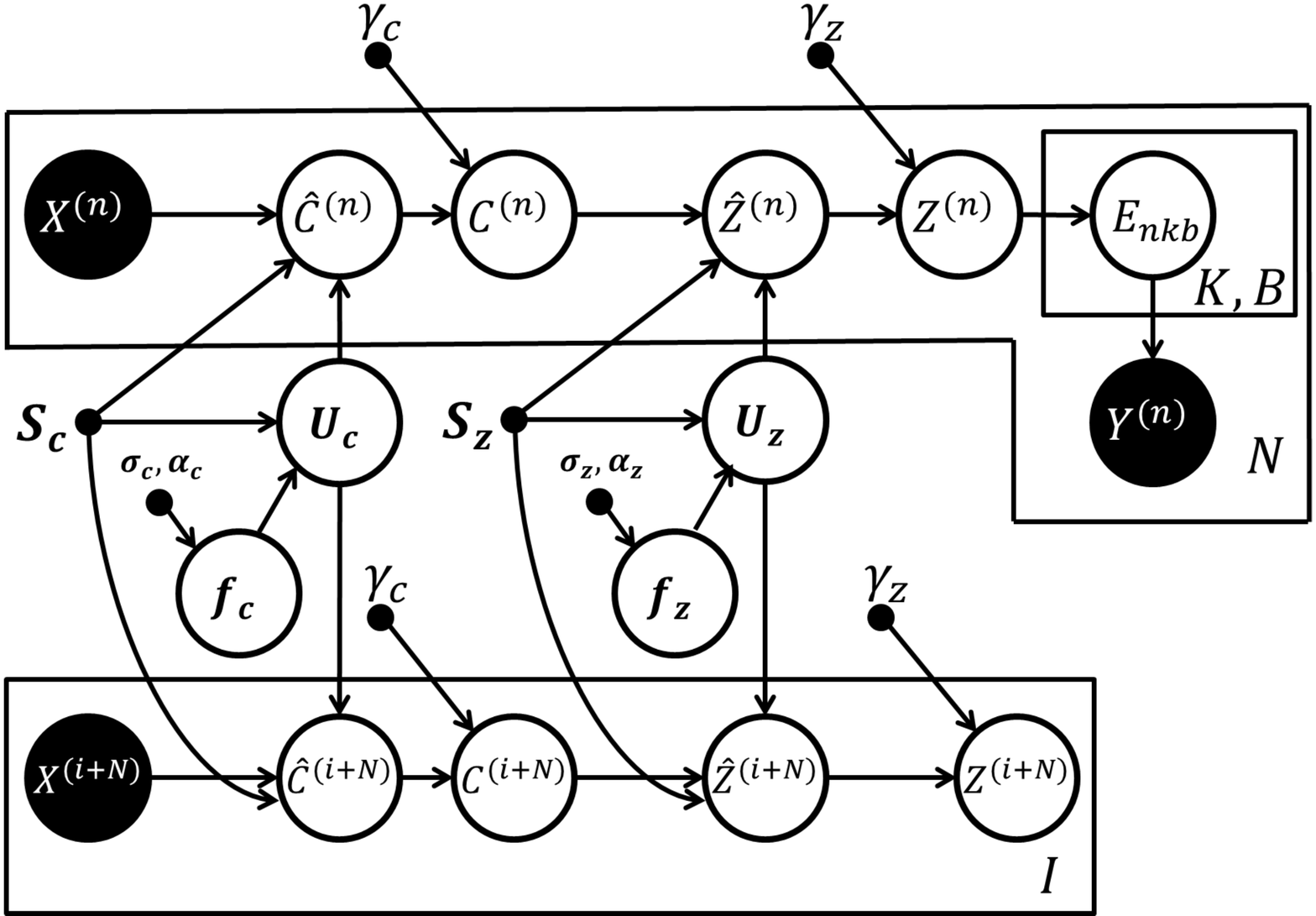}
\caption{A graphical model for the proposed method.}
\label{fig:pgm}
\end{figure}

In this section, we introduce the proposed method called Efficient Semi-supervised Multi-label Classification (ESMC). Fig. \ref{fig:pgm} illustrated the graphical model of the proposed method.

Given the feature vector $X^{(n)}$, the $\ell$-th dimension of the vector $\hat{C}^{(n)}$ - denoted by $\hat{c}^{(n)}_\ell$ - can be determined by the stochastic function $f_c^{(\ell)}$. Indeed, one may assume ${\hat{c}^{(n)}_\ell \approx f_c^{(\ell)}(X^{(n)})}$. However, as previously stated, using the traditional Gaussian processes \cite{gp} is prohibitive for large-scale datasets. In this regard, we use the idea of parameterizing the Gaussian processes by some pseudo-instances \cite{spgp}. 

To exploit the unlabeled data, as in SLRM \cite{slrm}, we seek to learn a smooth mapping to transform the feature vectors to the complete unobserved label vectors. In other words, let $X^{(j_1)}$ and $X^{(j_2)}$ be two feature vectors, where each of them can correspond to either a labeled or an unlabeled instance. According to Fig. \ref{fig:pgm}, our proposed model attempts to learn the stochastic functions in $\mathbf{f_c}$ and $\mathbf{f_z}$ with the desired smoothness property, that is if the vectors $X^{(j_1)}$ and $X^{(j_2)}$ are close to each other, the vectors $C^{(j_1)}$ and $C^{(j_2)}$, as well as the vectors $Z^{j_1}$ and $Z^{j_2}$, are most probably close to each other.

To handle missing labels, we used the EEOE approach of our previous work \cite{akpaper1}. The EEOE framework tackles the problem of missing labels by introducing a set of auxiliary random variables - referred to as \textit{experts} - which are denoted by $\lbrace E_{nkb} \rbrace _{n=1}^{N} {}_{k=1}^{K} {}_{b=1}^{B}$. 

The generative process of the proposed model is as follows:
\\1. Draw the stochastic functions $\mathbf{f_c} = \lbrace f_{c}^{(\ell)} \rbrace_{\ell=1}^{L}$:
\begin{equation}
f_{c}^{(\ell)} \sim \mathcal{GP}\big(\mathbf{0} \; , \; \kappa(\bullet,\bullet \; ; \sigma_c) \big).
\end{equation}
\\2. Draw the stochastic functions $\mathbf{f_z} = \lbrace f_{z}^{(k)} \rbrace_{k=1}^{K}$:
\begin{equation}
f_{z}^{(k)} \sim \mathcal{GP}\big(\mathbf{0} \; , \; \kappa(\bullet,\bullet \; ; \sigma_z) \big).
\end{equation}
\\3. Given the pseudo-instances, draw values of the functions in $\mathbf{f_c}$ at pseudo-instances:
\begin{equation}
for \; 1\le \ell \le L , 1\le m \le M_c:  u^{(\ell)}_{cm}\sim \mathcal{N} \big( \;f^{(\ell)}_c(s_c^{(m)}) , \alpha_c^2 \;\big).
\end{equation}
\\4. Given the pseudo-instances, draw values of the functions in $\mathbf{f_z}$ at pseudo-instances:
\begin{equation}
for \; 1\le k \le K, 1\le m \le M_z: u^{(k)}_{zm}\sim \mathcal{N} \big( \;f^{(k)}_z(s_z^{(m)}) , \alpha_z^2 \;\big).
\end{equation}
\\5. Given the representations in the feature space, generate the representations in the latent space:
\begin{equation}\label{eq:generate_hat_c}
\begin{split}
&for\;1\le n \le (N+I) \;\;,\;\; 1 \le \ell \le L:\\
&\hat{c}^{(n)}_\ell \sim \mathcal{N} 
\Big(
\kappa(X_\ell^{(n)},\mathbf{S_c})
\big[
\kappa(\mathbf{S_c},\mathbf{S_c})+\alpha_c^2 I_{{}_{M_c\times M_c}}
\big]^{-1} \times \\
&\;\;\;\;\;\;\;\;\;\;\;\;\;\;\;\;\big[u_{c1}^{(\ell)},...,u_{cM_c}^{(\ell)} \big]^T \;\;\;, \;\;\; \beta_c^2
 \Big).\\
\end{split}
\end{equation}
\\6. Add a Gaussian noise to the vector $\hat{C}^{(n)}$ and generate the vector $C^{(n)}$:
\begin{equation}\label{eq:generate_c}
\begin{split}
&for\; 1\le n \le N: \;\;C^{(n)} \sim \mathcal{N}\big(\hat{C}^{(n)} \;\;, \;\; \gamma_c^2I_{{}_{L\times L}} \big),\\
&for\; 1\le i \le I: \;\;C^{(N+i)} \sim \mathcal{N}\big(\hat{C}^{(N+i)} \;\;, \;\; \gamma_c^2I_{{}_{L\times L}} \big).\\
\end{split}
\end{equation}
\\7. Given the representations in the latent space, generate the vectors $\lbrace Z^{(n)}\rbrace_{n=1}^{N+I}$, where the $k$-th dimension of the vector $Z^{(n)}$ shows the suitability of the $k$-th label for the $n$-th instance:
\begin{equation}\label{eq:generate_hat_z}
\begin{split}
&for\;1\le n \le (N+I) \;\;,\;\; 1 \le k \le K:\\
&\hat{z}^{(n)}_k \sim \mathcal{N} 
\Big(
\kappa(C_\ell^{(n)},\mathbf{S_z})
\big[
\kappa(\mathbf{S_z},\mathbf{S_z})+\alpha_z^2 I_{{}_{M_z\times M_z}}
\big]^{-1} \times \\
&\;\;\;\;\;\;\;\;\;\;\;\;\;\;\;\;\big[u_{z1}^{(k)},...,u_{zM_z}^{(k)} \big]^T \;\;\;, \;\;\; \beta_z^2
 \Big).\\
\end{split}
\end{equation}
\\8. Add a Gaussian noise to the vector $\hat{Z}^{(n)}$ and generate the vector $Z^{(n)}$:
\begin{equation}\label{eq:generate_z}
\begin{split}
&for\; 1\le n \le N: \;\;Z^{(n)} \sim \mathcal{N}\big(\hat{Z}^{(n)} \;\;, \;\; \gamma_z^2I_{{}_{K\times K}} \big),\\
&for\; 1\le i \le I: \;\;Z^{(i+N)} \sim \mathcal{N}\big(\hat{Z}^{(i+N)} \;\;, \;\; \gamma_z^2I_{{}_{K\times K}} \big).\\
\end{split}
\end{equation}
\\9. For the $n$-th labeled instance:
\begin{itemize}
\item[]
a. Draw $B$ experts: 
\begin{equation}
for \; 1 \le b \le B: \;\;\; E_{nkb} \sim \mathcal{B}ernoulli \big( \sigma(\lambda z^{(n)}_k)),
\end{equation}
where $\lambda \in\mathbb{R}$ is a constant.
\item[]
b. Generate the label vector $Y^{(n)}$ deterministically:
\begin{equation}
for \; 1 \le k \le K: \;\;\; y^{(n)}_k  = E_{nk1} \frac{\sum_{b=1}^B E_{nkb}}{B}.
\end{equation}
\end{itemize}

In the above generative process, we do not generate $\hat{c}^{(n)}_\ell$ directly from the vector $X^{(n)}$ and the stochastic function $f_c^{(\ell)}$ using a traditional Gaussian process and instead we parameterize
the Gaussian process by some pseudo-instances. To this end, we firstly find the function $f_c^{(\ell)}$ in  pseudo samples as $\lbrace u_{cm}^{(\ell)} \rbrace _{\ell=1}^{L} {_{m=1}^{M_c}}$ where $u_{cm}^{(\ell)} \approx f_c^{(\ell)}(s_c^{(m)})$. Afterwards, we generate the random variable $\hat{c}^{(n)}_\ell$ by Eq. \ref{eq:generate_hat_c}. We use the same approach to parameterize the stochastic functions in $\mathbf{f_z}$.
Note that the addition of the Gaussian noise that is explained in Eq. \ref{eq:generate_c} and Eq. \ref{eq:generate_z} is only useful for statistical inference. Indeed, as explained in Eq. \ref{eq:generate_hat_c} and Eq. \ref{eq:generate_hat_z}, the uncertainty is modeled by drawing noisy samples from the stochastic functions. In other words, addition of this Gaussian noise, eliminates the terms 
$\mathbb{P}(C^{(n)}|X^{(n)})\times \mathbb{P}(Z^{(n)}|C^{(n)})$ in the likelihood equation, and makes 
the statistical inference feasible.
Accordingly, we set the parameters $\gamma_c$ and $\gamma_z$ to a small value.

\begin{figure}
\centering
\includegraphics[width=3.8in]{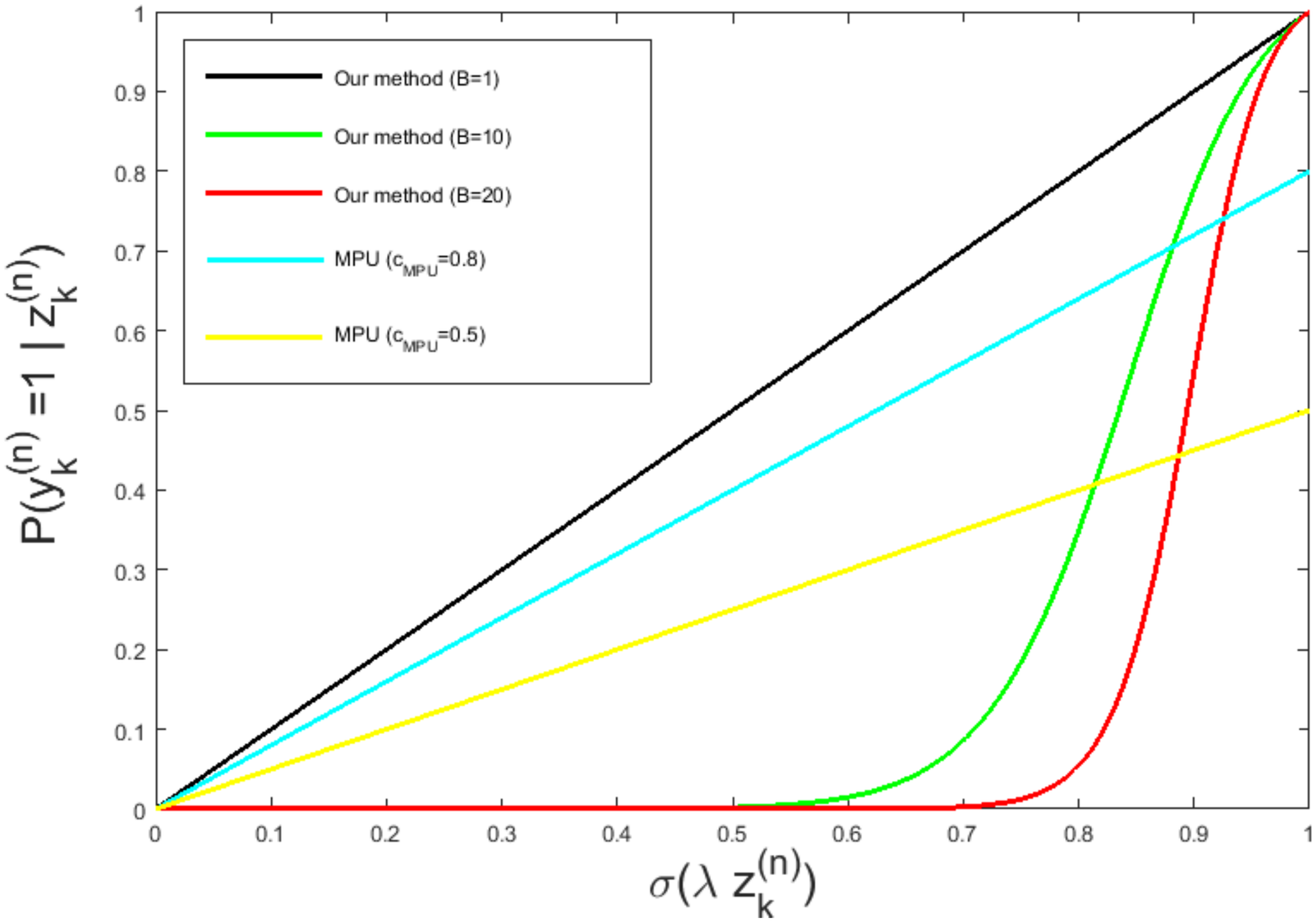}
\caption{Probability of assigning the $k$-th label to the $n$-th instance, versus 
the suitability of the label for the instance. The variable $c_{MPU}$ is the rate at which MPU \cite{mpu} randomly removes the unobserved complete label assignments to generate the observed label vectors.}
\label{fig:eeoe_framework}
\end{figure}

The effect of adding the experts (i.e. the random variables $\lbrace E_{nkb}\rbrace _{n=1}^{N} {}_{k=1}^{K} {}_{b=1}^{B}$) to handle missing labels is comprehensively explained in our previous article \cite{akpaper1}. It can be shown that after marginalizing out the experts, the following equation holds:

\begin{equation}\label{eq:eeoe_framework}
\mathbb{P}\big( y^{(n)}_k \; | z^{(n)}_k\; \big) \; = \;
\mathcal{B}ernoulli 
\Big(
\frac{\sigma(\lambda z^{(n)}_k)^B}{\sigma(\lambda z^{(n)}_k)^B + 1 - \sigma(\lambda z^{(n)}_k)}
\Big)
\end{equation}
Using Eq. \ref{eq:eeoe_framework}, Fig. \ref{fig:eeoe_framework} illustrates the probability of assigning the $k$-th label to the $n$-th instance, versus suitability of the label for the instance (i.e. $\sigma\big(\lambda z^{(n)}_k\big)$). According to Fig. \ref{fig:eeoe_framework}, when ${B>1}$, our proposed method assumes that with high probability, absolutely proper labels are not missed. However, MPU \cite{mpu} assumes that in presence of missing labels, even absolutely suitable label assignments have a little chance to be provided by the training set. In this regard, our mechanism for handling missing labels is preferable to that of MPU \cite{mpu}. Moreover, Fig. \ref{fig:eeoe_framework} demonstrates that, for instance when $B=20$, the model assigns a considerable chance to the case ${0.7 \le \sigma(\lambda z^{(n)}_k)\le 0.8}$ and $y^{(n)}_k=0$, and therefore is capable of handling missing labels.
\begin{figure}
\centering
\includegraphics[width=3.5in]{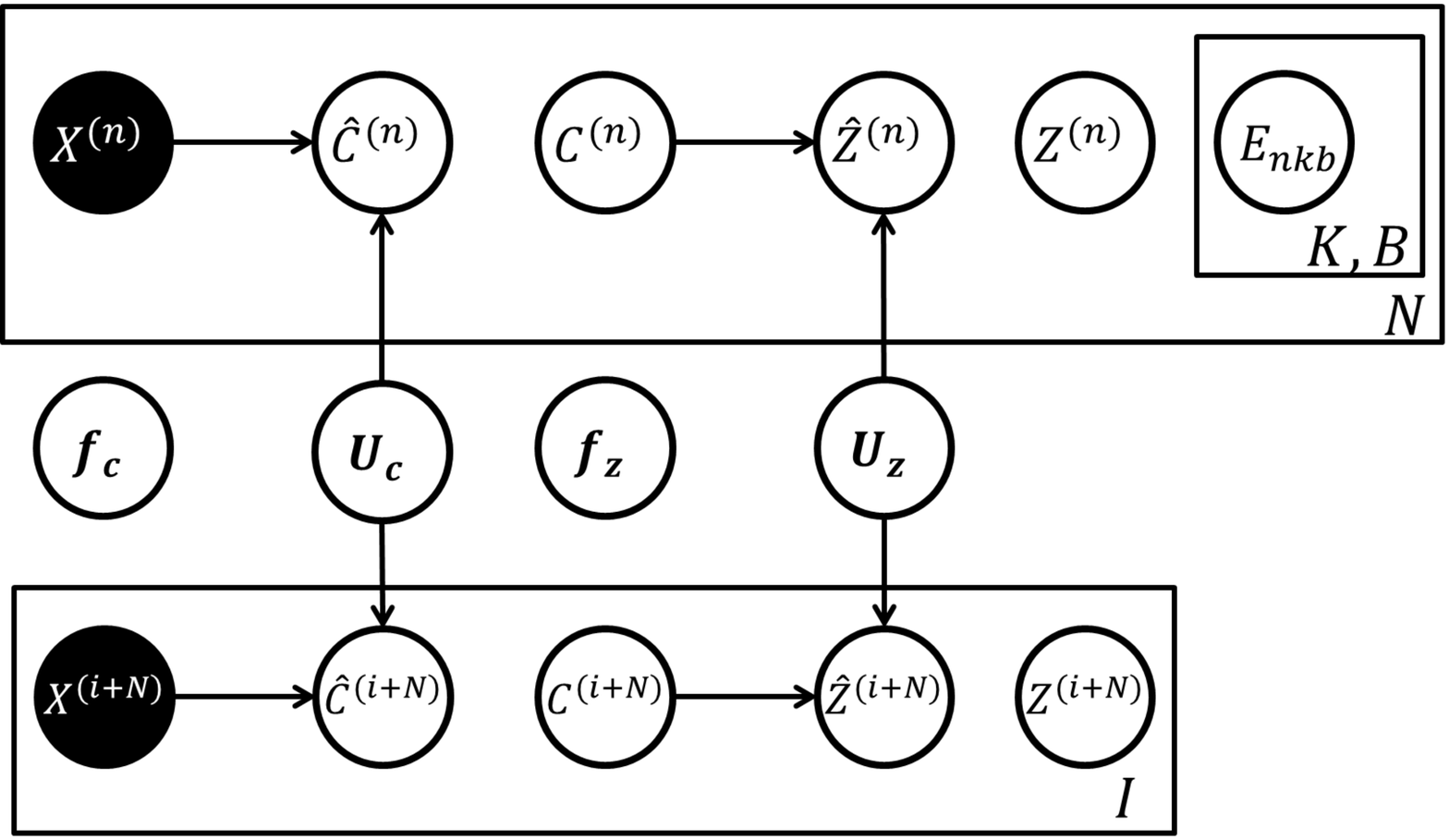}
\caption{We assume that the variational distribution can be factorized according to this Bayesian network.}
\label{fig:q_factorization}
\end{figure}
\subsection{Inference}
To predict labels for a test data $X_{test}$, we need to firstly predict the latent space representation of this instance, denoted by $C_{test}$, and then predict the label vector $Y_{test}$. To this end, we seek to find the distribution 
$\small{\mathbb{P}\big(\lbrace f_c^{(\ell)}\rbrace)_{\ell=1}^{L}\;,\; \lbrace f_z^{(k)}\rbrace_{k=1}^{K}\; | \; \lbrace X^{(n)}\rbrace_{n=1}^{(N+I)} 
,\lbrace Y^{(n)}\rbrace_{n=1}^{N}\big)}$ in order to predict the label vector $Y_{test}$ as
$Y_{test} = E\big[\mathbf{f_z}\big(E[\mathbf{f_c}(X_{test})\big)]  \big]$. However, for the proposed probabilistic model, finding the posterior distribution is intractable. Moreover, in this model, because of existence of some factors like $\mathbb{P}(\hat{C}^{(n)}\;|\; X^{(n)},\mathbf{U_c})\times \mathbb{P}(C^{(n)}\;|\;\hat{C}^{(n)})$, the mean-field variational inference is intractable. In this regard, we adapted the evidence lower bound which has also been used in the methods VAE-DGP \cite{vadgp}, DGP \cite{dgp}, as well as GP-LVM \cite{gplvm}. Note that because of introducing the auxiliary random variables (i.e. experts), we make additional assumptions to adapt this lower bound to our probabilistic model. More precisely, we make the following assumptions about the variational distribution:
\\1. We assume that the variational distribution $q$ can be factorized according to the Bayesian network of Fig. \ref{fig:q_factorization}. 

In other words, we consider the family of distributions that can be factorized as follows:
\begin{equation}
\begin{split}
&q = 
\Big[\prod_{\ell=1}^{L} q\big(f_c^{(\ell)}\big) \Big]
\times \Big[\prod_{k=1}^{K} q\big(f_z^{(k)}\big) \Big]\times 
\Big[ \prod_{m=1}^{M_c} \prod_{\ell=1}^{L}q\big( u_{cm}^{(\ell)}\big) \Big] \times\\
&\;\;\;\;\;\;\,\Big[ \prod_{m=1}^{M_z} \prod_{k=1}^{K}q\big( u_{zm}^{(k)}\big) \Big] \times
\Big[\prod_{n=1}^{N+I} q\big(\hat{C}^{(n)}|\mathbf{U_c},X^{(n)} \big)\Big] \times\\
&\;\;\;\;\;\;\,\Big[\prod_{n=1}^{N+I}q\big( \hat{Z}^{(n)}| C^{(n)},\mathbf{U_z}\big)\Big] \times
\Big[\prod_{n=1}^{N+I} q\big(C^{(n)}\big)\Big]\times\\
&\;\;\;\;\;\;\,\Big[\prod_{n=1}^{N+I} q\big(Z^{(n)}\big)\Big]\times
\Big[ \prod_{n=1}^{N} \prod_{k=1}^{K} \prod_{b=1}^{B} q\big( E_{nkb}\big)\Big].
\end{split}
\end{equation}
By doing so, it is still impractical to optimize the evidence lower bound. Thus, we make the following assumption too.
\\2. We suppose that the following equations hold:
\begin{equation}\label{eq:assumption_2_1}
q\big(\hat{C}^{(n)}|\mathbf{U_c},X^{(n)} \big) = \mathbb{P}\big(\hat{C}^{(n)}|\mathbf{U_c},X^{(n)} \big)
\end{equation}
\begin{equation}\label{eq:assumption_2_2}
q\big(\hat{Z}^{(n)}|\mathbf{U_z},C^{(n)} \big) = \mathbb{P}\big(\hat{Z}^{(n)}|\mathbf{U_z},C^{(n)} \big),
\end{equation}
where the right hind side of Eq. \ref{eq:assumption_2_1} and \ref{eq:assumption_2_2} corresponds to the conditional distributions which are explained in Eq. \ref{eq:generate_hat_c}.

These two assumptions make it feasible to maximize the evidence lower bound. More precisely, let $\mathcal{L}(q)$ - the evidence lower bound - be defined as follows:
\begin{equation}\label{eq:elbo_definition}
\mathcal{L}(q) \triangleq \int q\;ln\big( \frac{\mathbb{P}}{q}\big)\; d(the\;latent\;variables). 
\end{equation}  
By assuming that Eq. \ref{eq:assumption_2_1} and \ref{eq:assumption_2_2} hold, one can omit the factors $\mathbb{P}\big(\hat{C}^{(n)}|\mathbf{U_c},X^{(n)} \big)$ and $\mathbb{P}\big(\hat{Z}^{(n)}|\mathbf{U_z},C^{(n)} \big)$ from the numerator, and the factors $q\big(\hat{C}^{(n)}|\mathbf{U_c},X^{(n)} \big)$ and $q\big(\hat{Z}^{(n)}|\mathbf{U_z},C^{(n)} \big)$ from the denominator of the term $\frac{\mathbb{P}}{q}$, in the right hand side of Eq. \ref{eq:elbo_definition}.
Nonetheless, because of using the sigmoid function in the generative process of our model, it is not straightforward to maximize the evidence lower bound. In this regard, we approximate the sigmoid function $\sigma(t)$ by the function $G(t,\xi)$ \cite{sigmoid_approx}, which is defined as follows:
\begin{equation}\label{eq:define_g}
G(t,\xi) \triangleq \sigma(\xi)exp\lbrace \frac{t-\xi}{2} - \frac{tanh(\frac{\xi}{2})}{4\xi} (t^2 - \xi^2)\rbrace.
\end{equation}
More precisely, we approximate the terms ${\mathbb{P}(E_{nkb}|z^{(n)}_k)}$ as follows:
\begin{equation}\label{eq:enkb_approximation}
\begin{split}
p(E_{{}_{nkb}}|z^{(n)}_k) &= \sigma(
\lambda z^{(n)}_k)^{E_{{}_{nkb}}} \big(1-\sigma(
\lambda z^{(n)}_k)\big)^{1-E_{{}_{nkb}}}\\
&= \big(\frac{\sigma(\lambda z^{(n)}_k)}{1-\sigma(\lambda z^{(n)}_k)} \big)^{E_{nkb}} \; \big(1-\sigma(\lambda z^{(n)}_k)\big)\\
&= exp\big( E_{{}_{nkb}}\lambda z^{(n)}_k \big)\;\sigma\big( -\lambda z^{(n)}_k\big)\\
&\approx exp\big( E_{{}_{nkb}}\lambda z^{(n)}_k \big)\; G\big( -\lambda z^{(n)}_k \;,\; \xi_{nk}\big).
\end{split}
\end{equation}
Note that the parameters $\lbrace \xi_{nk}\rbrace_{n=1}^{N} {}_{k=1}^{K}$ are auxiliary variational parameters. One can set $\frac{\partial \mathcal{L}(q)}{\partial \xi_{nk}}$ to zero and get: 
\begin{equation}
\xi_{nk} = \pm \lambda \;E_{{}_{Z^{(n)}\sim q(Z^{(n)})}}\big[z^{(n)}_k \big].
\end{equation} 
By making the first and the second assumption and using the approximation explained in Eq. \ref{eq:enkb_approximation}, maximizing the evidence lower bound is straightforward. 

Suppose that some vectors in the latent space denoted by $\lbrace C^{(j_1)},...,C^{(j_t)}\rbrace$ are very close to the vector $C^{(n)}$. Furthermore, assume that ${y^{(n)}_k=1\;,\;y^{(j_1)}_k=0,...,\;y^{(j_t)}_k=0}$. As a consequence, the posterior distribution of the proposed model tend to assign a considerable chance to the case $z^{(n)}_k\ll 0$, and so neglects the assignment of the $k$-th label to the $n$-th instance. Indeed, although the terms $\lbrace \mathbb{P}(E_{nkb}|z^{(n)}_k)\rbrace_{b=1}^B$ might lead to assigning a considerable chance to the case $0 \ll z^{(n)}_k$, the terms $\lbrace\mathbb{P}(\hat{z}^{(n)}_k\;|\;C^{(n)},\mathbf{U_z})\rbrace$ might lead to the case $z^{(n)}_k\ll 0$ to learn a smoother mean for the stochastic function $f^{(k)}_z(\bullet)$. To avoid this problem, if $y^{(n)}_k=1$, we set the $k$-th dimension of the mean of the variational distribution $q\big(Z^{(n)}\big)$ to a large number.

The pseudo-instances are not constrained to be a subset of the data \cite{spgp}. However, for simplicity, we set the pseudo-instances $\lbrace s^{(m)}_c \rbrace_{m=1}^{M_c}$ to be a subset of the feature vectors (i.e. $\lbrace X^{(n)} \rbrace_{n=1}^{N+I}$).
Moreover, in each iteration of the variational inference, we set $M_c = M_z$ and update the pseudo-instances $\lbrace s^{(m)}_z \rbrace_{m=1}^{M_z}$ as follows:
\begin{equation}
s^{(m)}_z \leftarrow E \big[\mathbf{f_c}(s_c^{(m)}) \big]
\end{equation}

\section{Experiments}
\subsection{Experimental Setup}
To validate the performance of the proposed method, we conducted experiments on seven real-world datasets, \textit{corel5k}, \textit{iaprtc12} and 
\textit{espgame} from LEAR website \cite{lear_datasets}; and \textit{CAL500}, \textit{NUS-WIDE}, \textit{mediamill} and \textit{delicious} from Mulan \cite{mulan}. Some statistics of these datasets is provided in Tab. \ref{table:datasets_stat}.
\begin{table}
\caption{Statistics of the used datasets.}
\label{table:datasets_stat}
\begin{tabular}{c?c|c|c|c |c}
 \textbf{Dataset}& ${N}$ & $F$ & $K$ & {\# test instances} &{ \# avg label per sample}\\
\hline
CAL500 & 400 & 68 & 174 & 102&26.044\\
corel5k & 3999 & 100 & 260 & 1000&3.3883\\
iaprtc12 & 15701 & 100 & 291 & 3926&5.6990\\
espgame & 16616 & 100 &268 & 4154&4.6741\\
delicious & 12920 & 983 &500 & 3185&19.0213\\
mediamill & 30993 & 120 &101 & 1291&4.3784\\
nus-wide & 161789 & 500 &81 & 107859&1.8655\\
\end{tabular}
\end{table}
We compared the performance of our ESMC method to that of some previous methods, including SLRM \cite{slrm}, LEML \cite{leml},
 FaIE \cite{faie} and FastTag \cite{fasttag}. We used the implementation of FastTag \cite{fasttag,fasttag_code}, provided by the authors of the article. Moreover, we implemented SLRM \cite{slrm} and FaIE \cite{faie}. The parameters of FastTag \cite{fasttag} are set to default values in its code. 
In all of the experiments, according to the article of FaIE \cite{faie}, the parameter $\alpha_{{}{{}_{FaIE}}}$ in FaIE \cite{faie} is selected from the set 
$\lbrace 10^{-1} , 10^0,...,10^4\rbrace$.
Similarly, the parameters $\lambda_{{}_{{}_{SLRM}}}$ and $\gamma_{{}_{{}_{SLRM}}}$ of SLRM \cite{slrm} are selected from the set $\lbrace 10^{-3},...,10^{3}\rbrace$ as suggested in \cite{slrm}.
Since SLRM \cite{slrm} and FaIE \cite{faie} are not fast methods, using cross validation to set these parameters is time consuming, specially when one wants to evaluate these methods in many different settings. In this regard, instead of using cross validation, in all of the experiments we selected values for the parameters of FaIE \cite{faie} and SLRM \cite{slrm} from the above sets that lead to the best prediction performance (in terms of the maximum achievable micro-F1 \cite{survey_1}, when using different thresholds to convert the continuous predictions to binary label vectors of length $K$) on the test instances. It is clearly above the prediction performance obtained for these methods by setting parameters using cross-validation.
We set the kernel parameter $\sigma_c$ of our method as twice the mean Euclidean distance between feature vectors \cite{faie}.
The kernel parameter of FaIE \cite{faie} has also been set similarly.
Moreover, the parameter $B$ (i.e. number of experts) is set as follows:
\begin{equation}\label{eq:select_b}
B = min\lbrace \frac{\text{\# zero elements of the matrix } \mathbf{Y}}{\text{\# one elements of the matrix } \mathbf{Y}} \; ,\; 100\rbrace
\end{equation}
The number of the pseudo-instances is set as:
\begin{equation}
M_c = M_z = 
\begin{cases}
0.1\times N\;\;\;\;\;\;\;\;\;\;&1\le N < 10000\\
0.01 \times N\;\;\;\;\;\;\;\; &10000 \le N < 20000\\
400\;\;\;\;\;\;\;\;\; &20000< N\\
\end{cases}
\end{equation}
Increasing the number of pseudo instances can improve results of our method. However, the time cost of our method is affected too.

In the test phase, SLRM \cite{slrm}, LEML \cite{leml},
 FaIE \cite{faie}, FastTag \cite{fasttag}, and the proposed ESMC method produce a real-valued label space representation. Transforming these real-valued vectors to binary label vectors is a challenging task which dramatically affects the prediction performance. 
Indeed, evaluating the above methods by their produced binary label vectors may lead to unfair evaluations. In this regard, to avoid the problem of producing binary label vectors, we used three rank-based evaluation metrics that are Area Under ROC Curve (AUC) \cite{survey_1}, 
coverage \cite{survey_1}, and Precision@k \cite{leml}. These metrics are widely used for evaluating multi-label classifiers. To evaluate the performance of the methods
on a dataset, we randomly partitioned its instances three times
and averaged the performance obtained over these runs. When
partitioning the dataset into the training and the test set, we determined the number of training instances according to Tab. \ref{table:datasets_stat}.

\subsection{Experimental Results}

\begin{figure}[t]
\begin{subfigure}{.25\textwidth}
  \centering
 \includegraphics[width=1.05\linewidth]{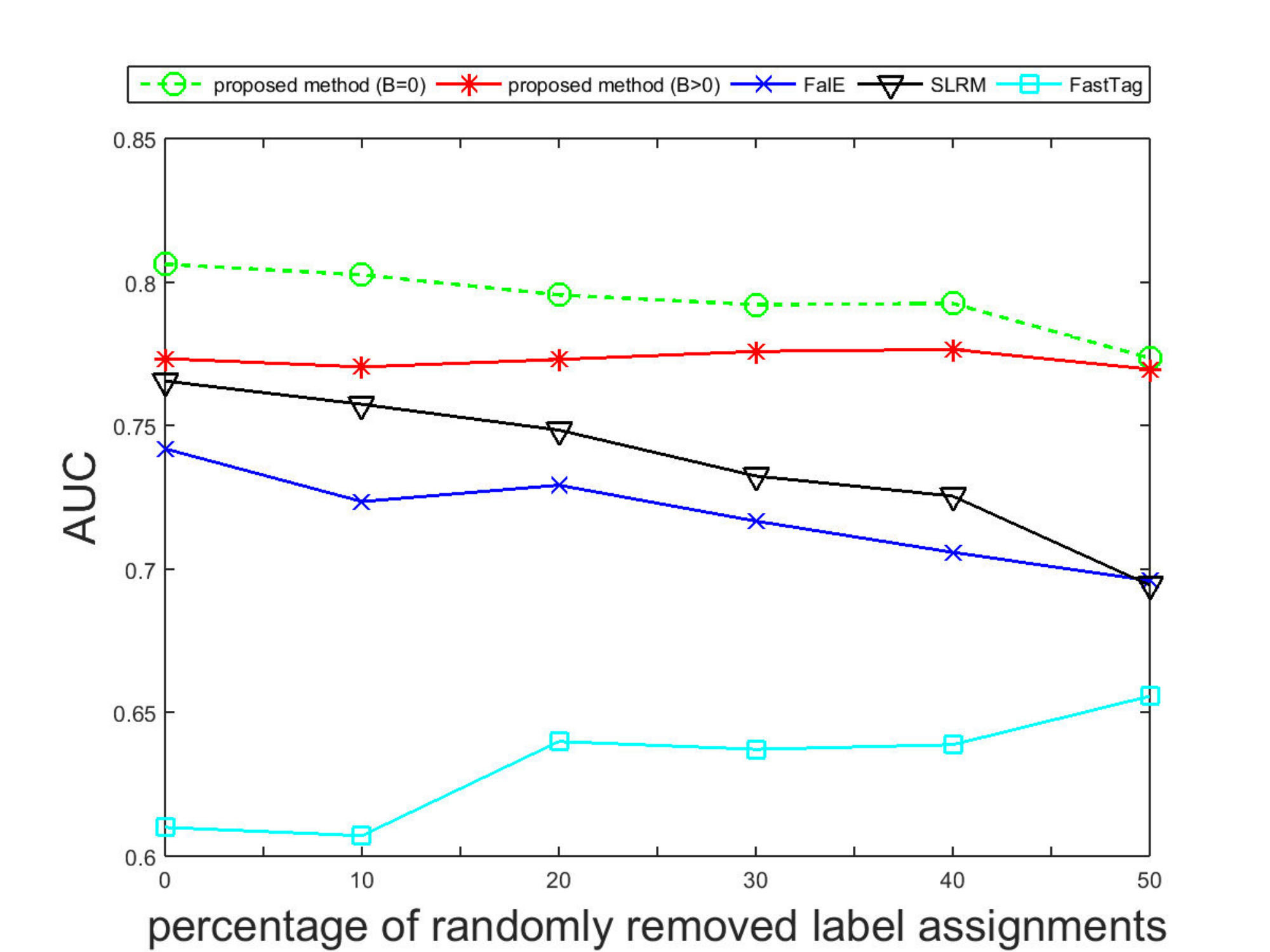}
  \caption*{(a) AUC}
\end{subfigure}%
\begin{subfigure}{.25\textwidth}
  \centering
  \includegraphics[width=1.05\linewidth]{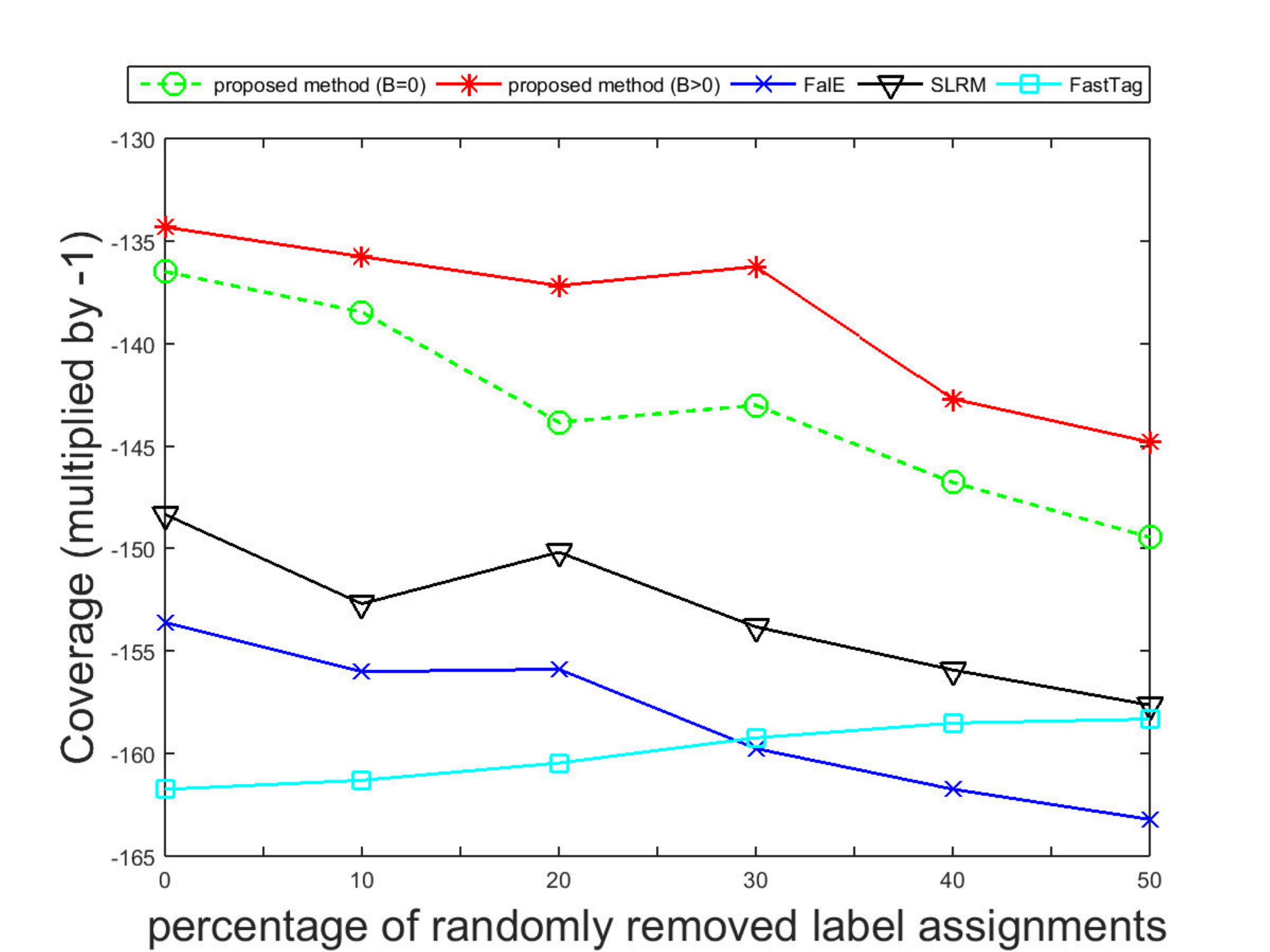}
  \caption*{(b) -Coverage}
\end{subfigure}
\caption{{Results on CAL500, for different percentages of randomly removed label assignments (i.e. simulating missing labels).}}
\label{fig:cal500_removy}
\end{figure}

\begin{figure}[t]
\begin{subfigure}{.25\textwidth}
  \centering
 \includegraphics[width=1.05\linewidth]{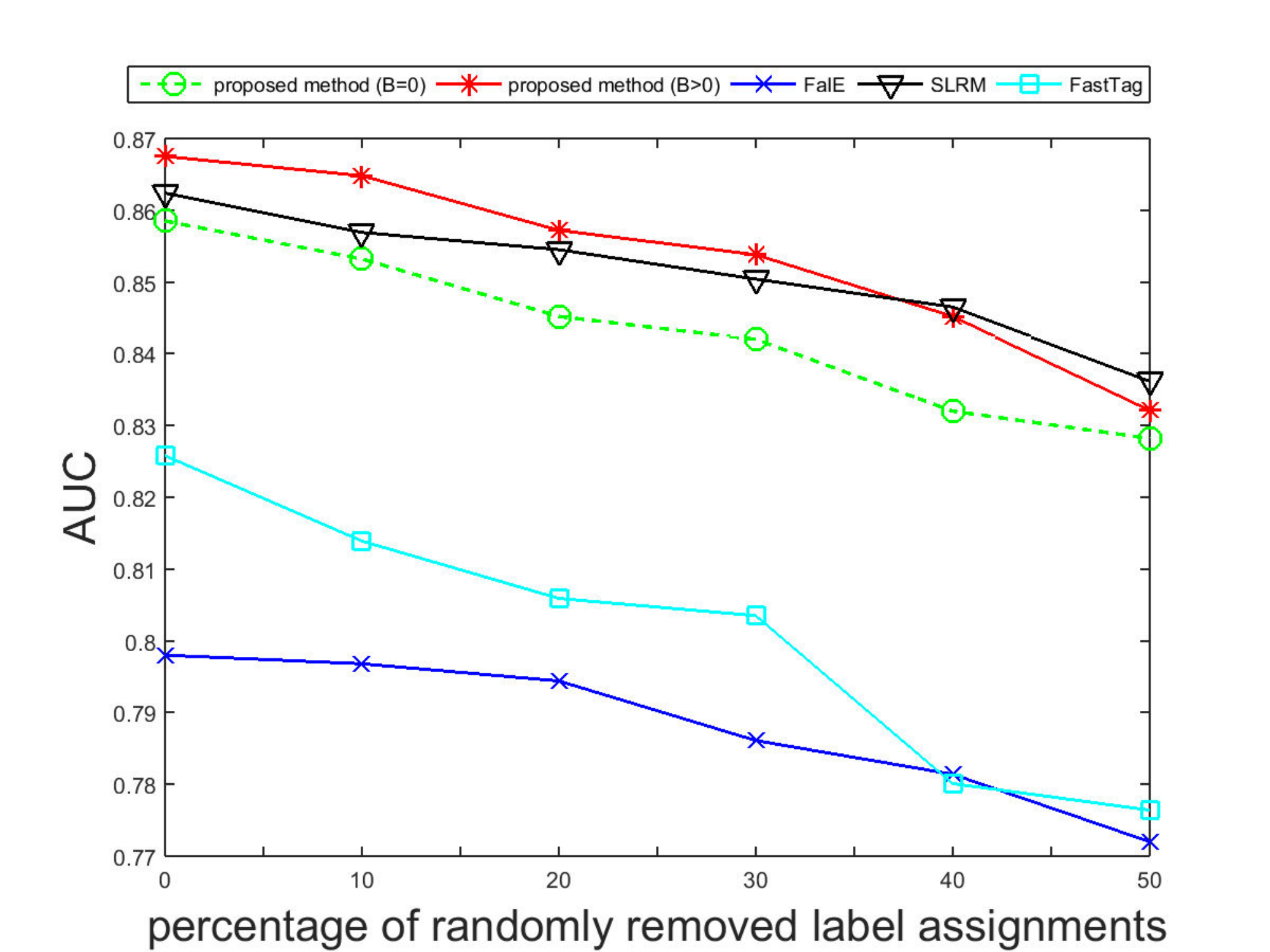}
  \caption*{(a) AUC}
  \label{fig:sfig1}
\end{subfigure}%
\begin{subfigure}{.25\textwidth}
  \centering
  \includegraphics[width=1.05\linewidth]{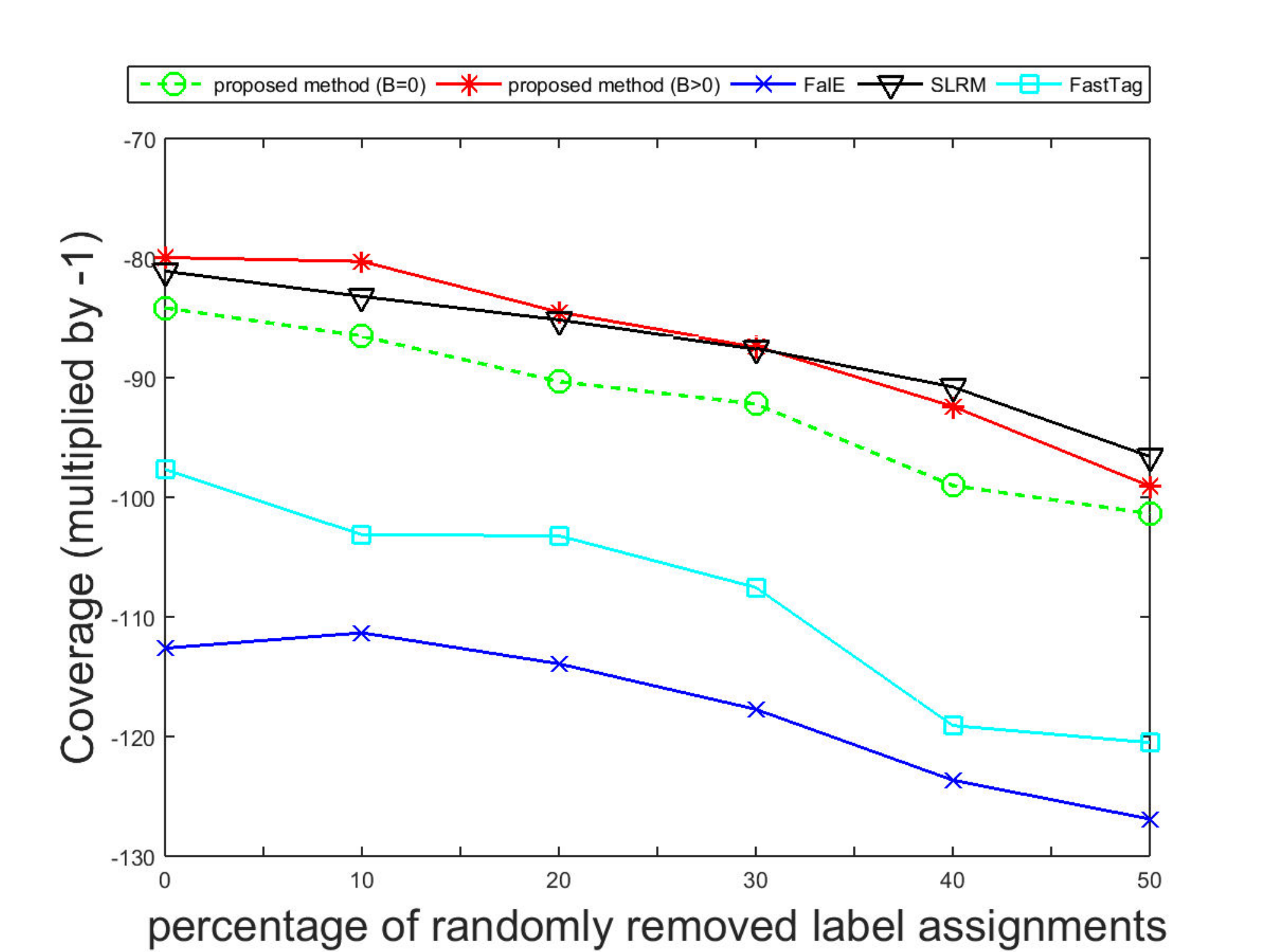}
  \caption*{(b) -Coverage}
  \label{fig:sfig2}
\end{subfigure}
\caption{{Results on corel5k, for different percentages of randomly removed label assignments (i.e. simulating missing labels).}}
\label{fig:corel5k_removy}
\end{figure}

\begin{figure}[t]
\begin{subfigure}{.25\textwidth}
  \centering
 \includegraphics[width=1.05\linewidth]{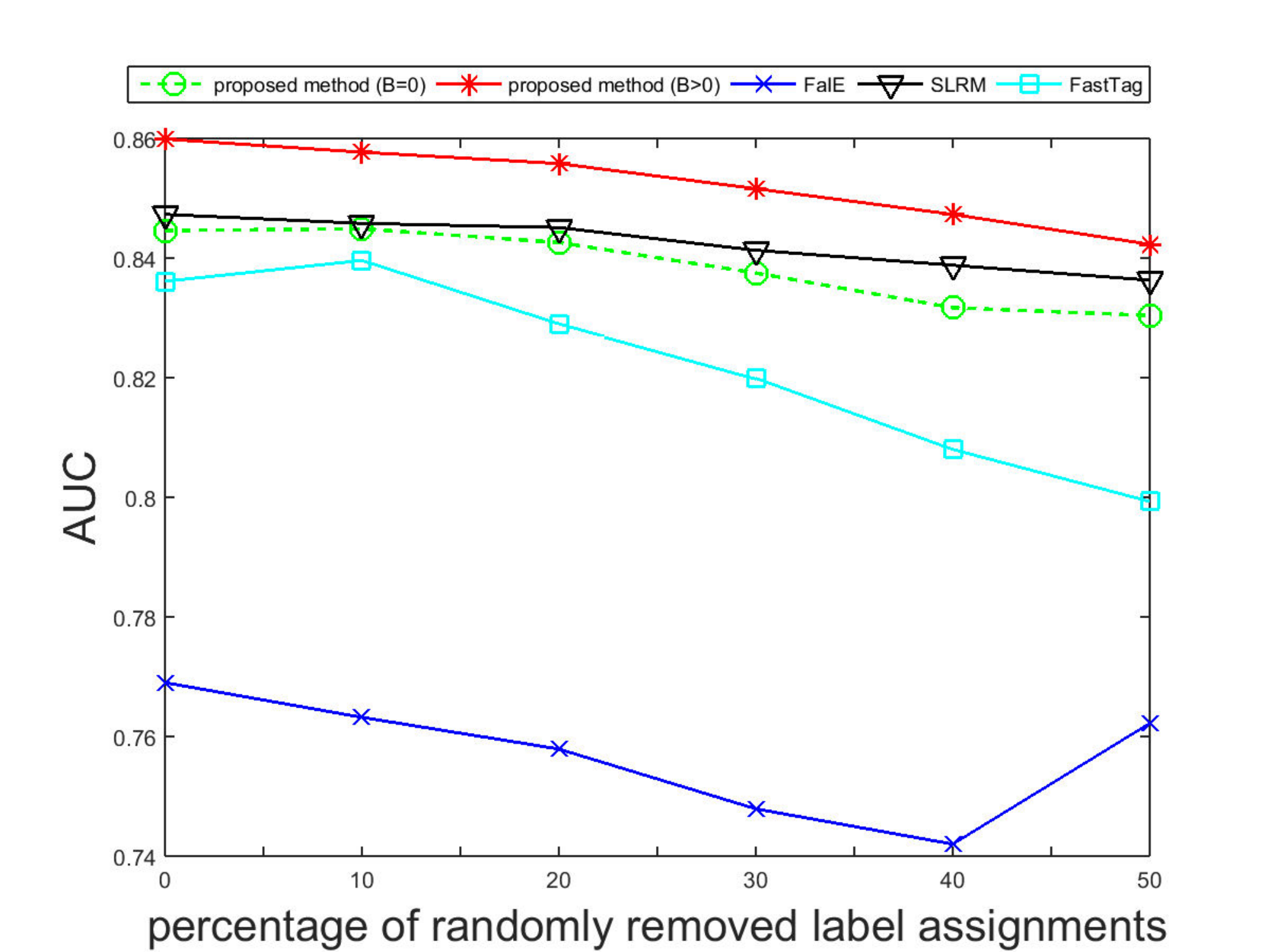}
  \caption*{(a) AUC}
  \label{fig:sfig1}
\end{subfigure}%
\begin{subfigure}{.25\textwidth}
  \centering
  \includegraphics[width=1.05\linewidth]{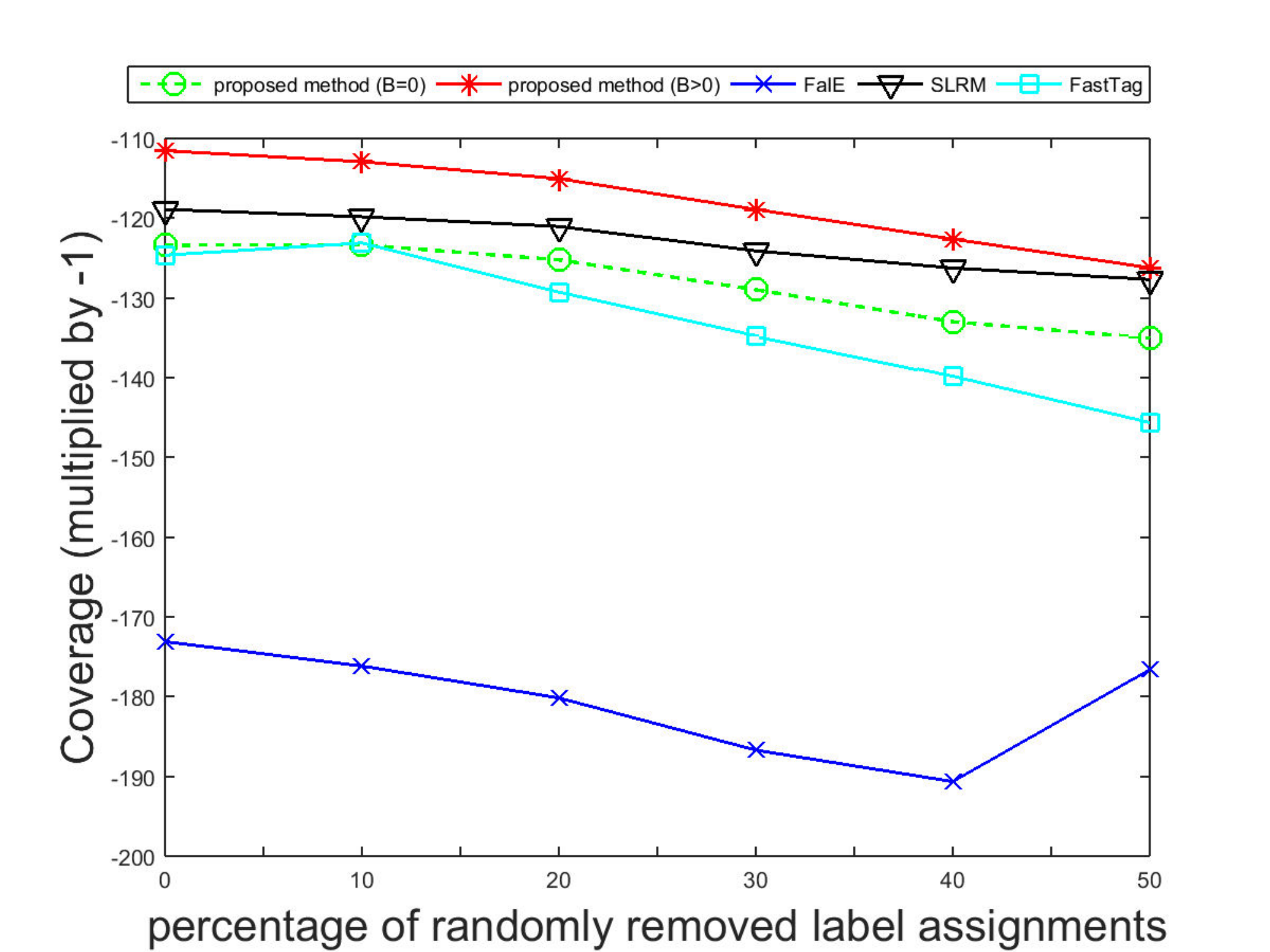}
  \caption*{(b) -Coverage}
  \label{fig:sfig2}
\end{subfigure}
\caption{{Results on iaprtc12, for different percentages of randomly removed label assignments (i.e. simulating missing labels).}}
\label{fig:iaprtc12_removy}
\end{figure}

\begin{figure}[t]
\begin{subfigure}{.25\textwidth}
  \centering
 \includegraphics[width=1.05\linewidth]{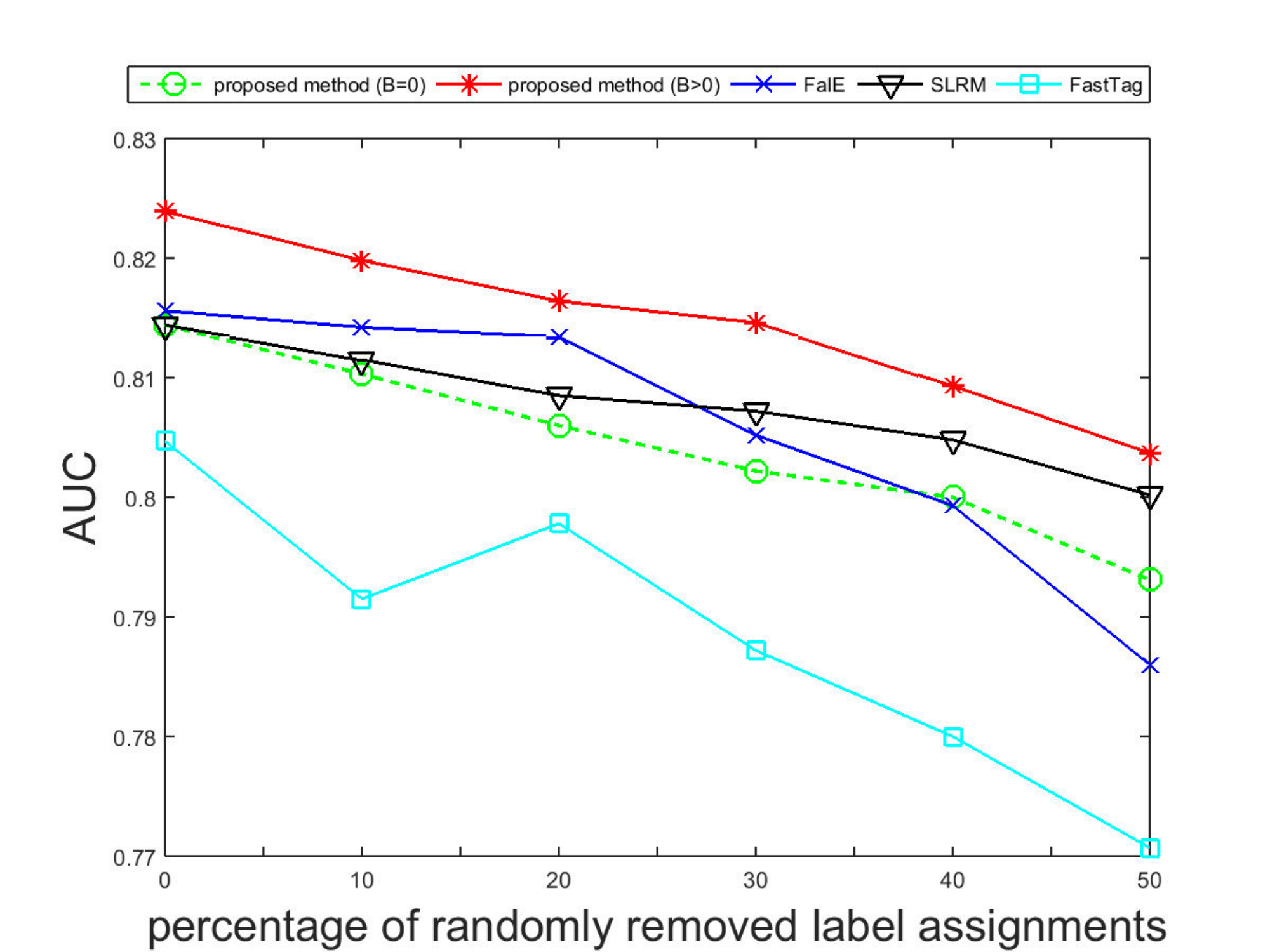}
  \caption*{(a) AUC}
  \label{fig:sfig1}
\end{subfigure}%
\begin{subfigure}{.25\textwidth}
  \centering
  \includegraphics[width=1.05\linewidth]{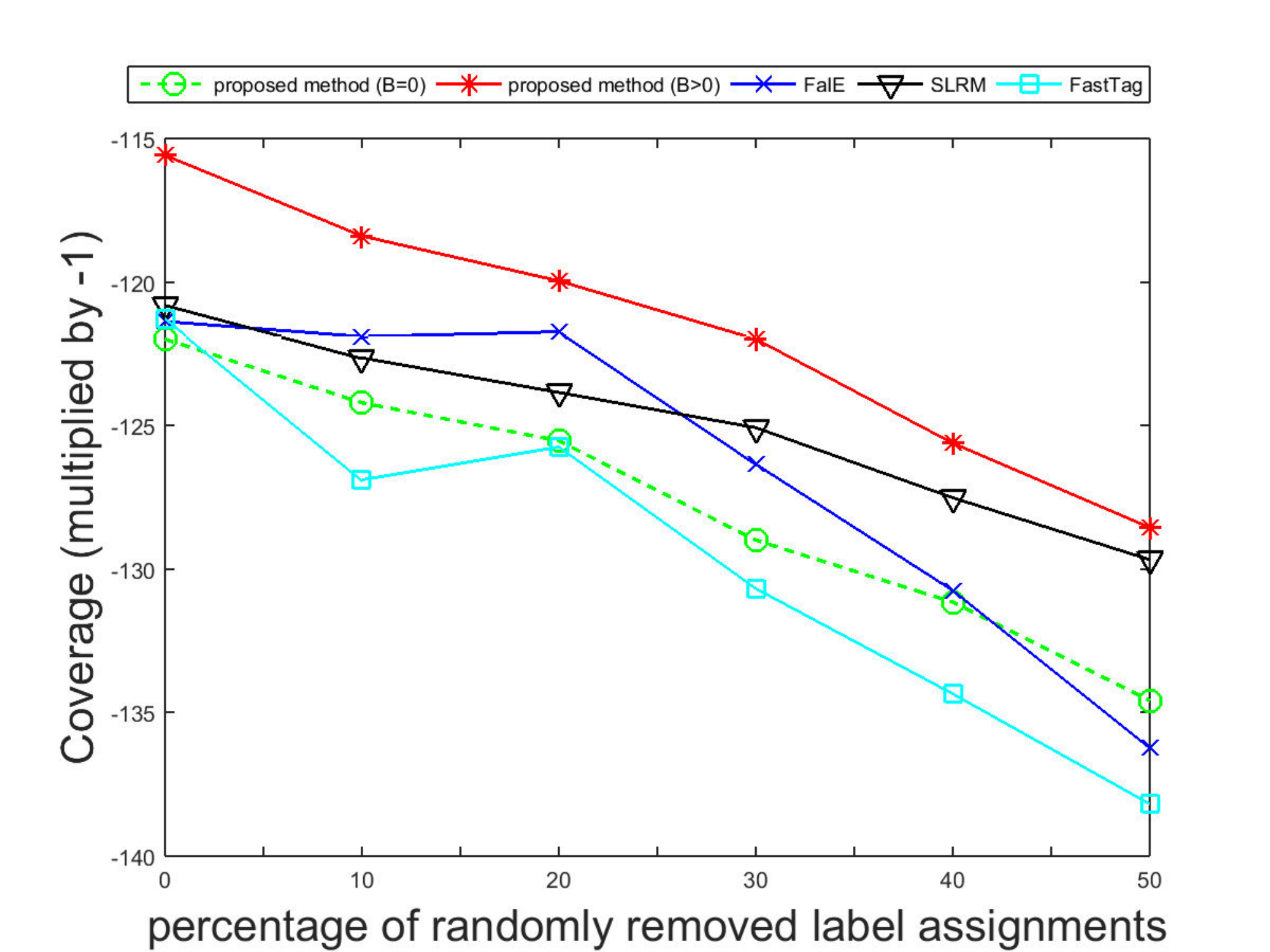}
  \caption*{(b) -Coverage}
  \label{fig:sfig2}
\end{subfigure}
\caption{{Results on espgame, for different percentages of randomly removed label assignments (i.e. simulating missing labels).}}
\label{fig:espgame_removy}
\end{figure}

\begin{figure}[t]
\begin{subfigure}{.25\textwidth}
  \centering
 \includegraphics[width=1.05\linewidth]{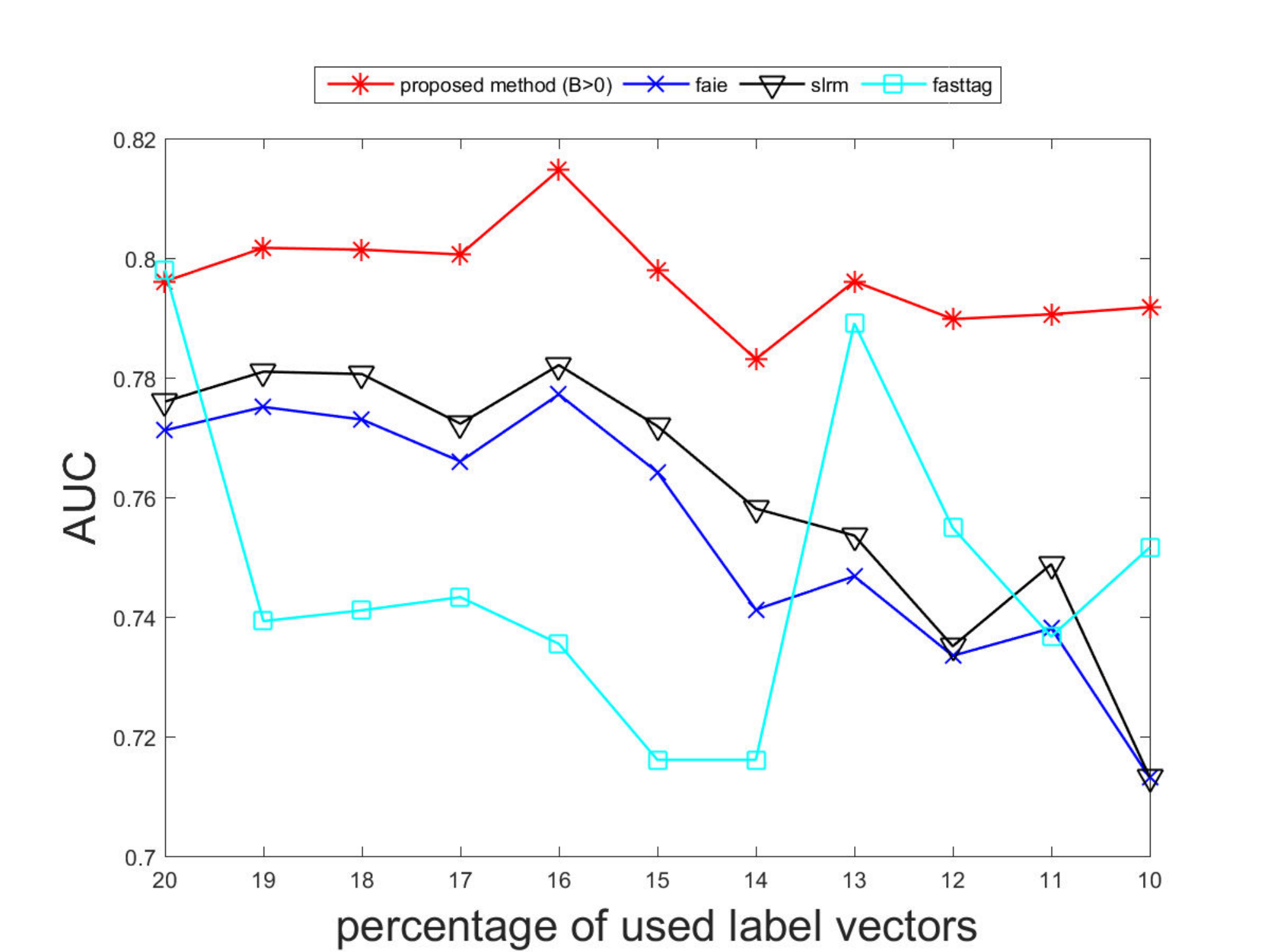}
  \caption*{(a) AUC}
  \label{fig:sfig1}
\end{subfigure}%
\begin{subfigure}{.25\textwidth}
  \centering
  \includegraphics[width=1.05\linewidth]{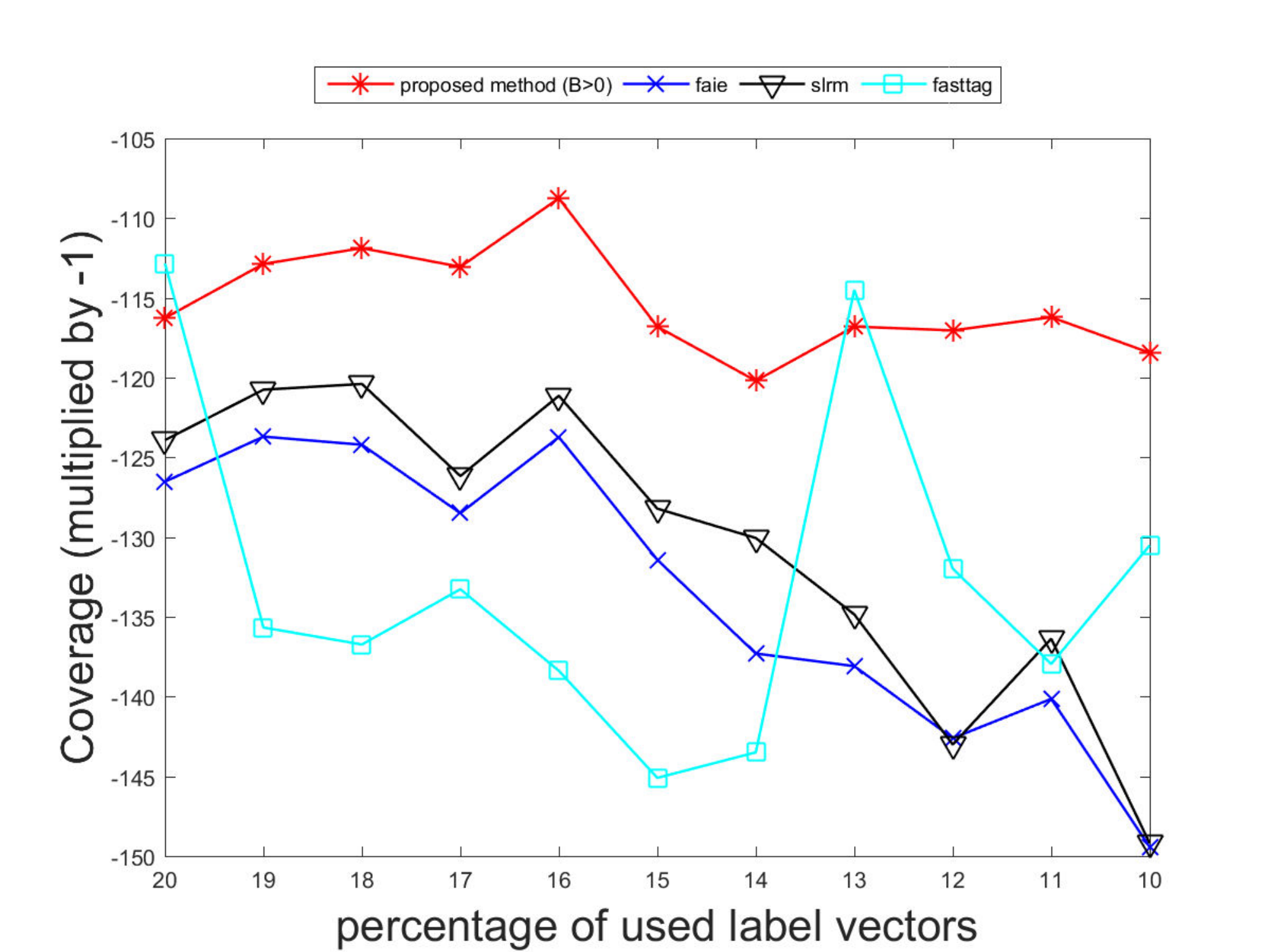}
  \caption*{(b) -Coverage}
  \label{fig:sfig2}
\end{subfigure}
\caption{Results on corel5k, for different percentages of randomly selected label vectors (i.e. simulating the semi-supervised setting).}
\label{fig:corel5k_semisupervised}
\end{figure}

\begin{figure}[t]
\begin{subfigure}{.25\textwidth}
  \centering
 \includegraphics[width=1.05\linewidth]{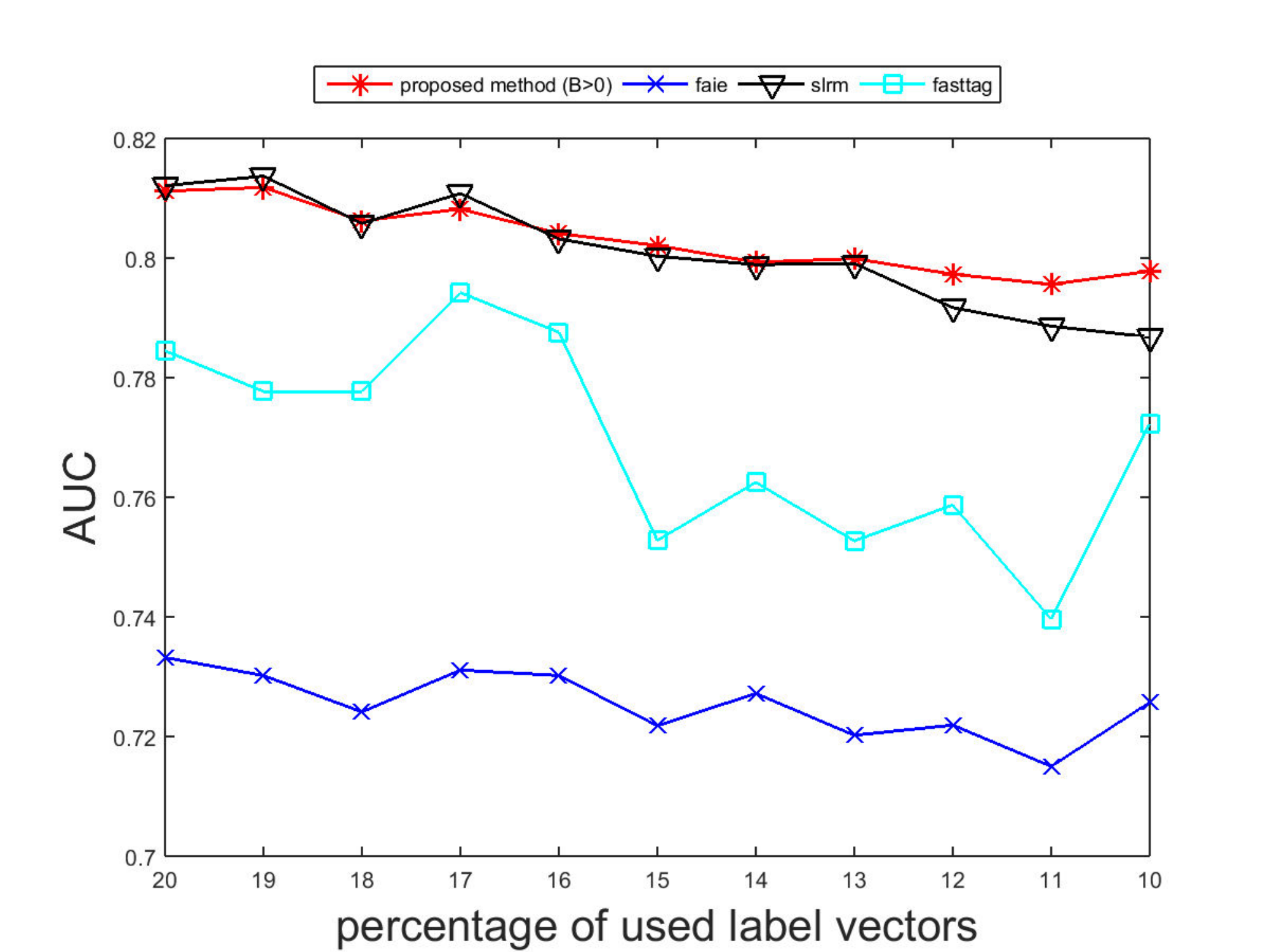}
  \caption*{(a) AUC}
  \label{fig:sfig1}
\end{subfigure}%
\begin{subfigure}{.25\textwidth}
  \centering
  \includegraphics[width=1.05\linewidth]{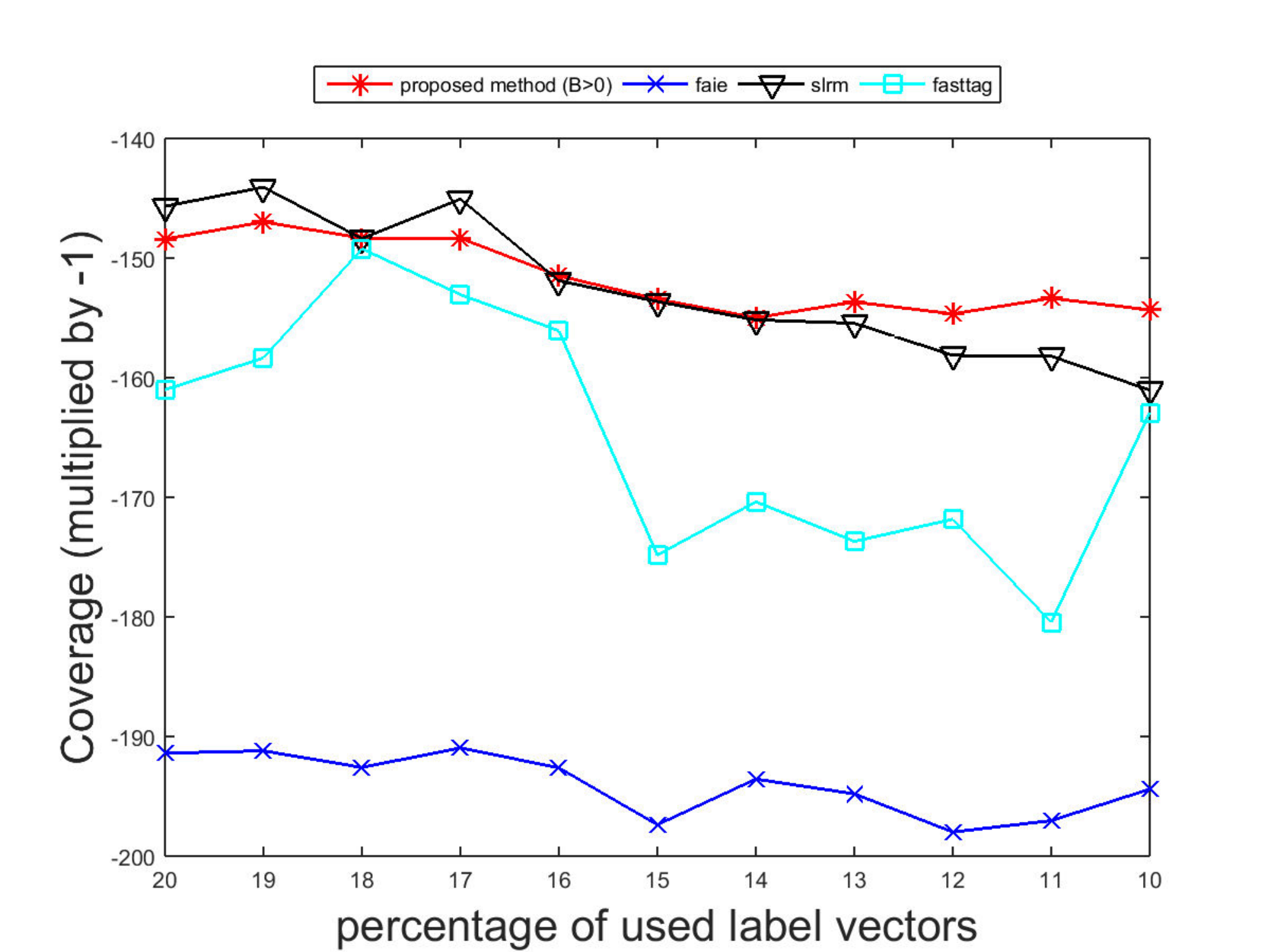}
  \caption*{(b) -Coverage}
  \label{fig:sfig2}
\end{subfigure}
\caption{{Results on iaprtc12, for different percentages of randomly selected label vectors (i.e. simulating the semi-supervised setting).}}
\label{fig:iaprtc12_semisupervised}
\end{figure}

\begin{figure}[t]
\begin{subfigure}{.25\textwidth}
  \centering
 \includegraphics[width=1.05\linewidth]{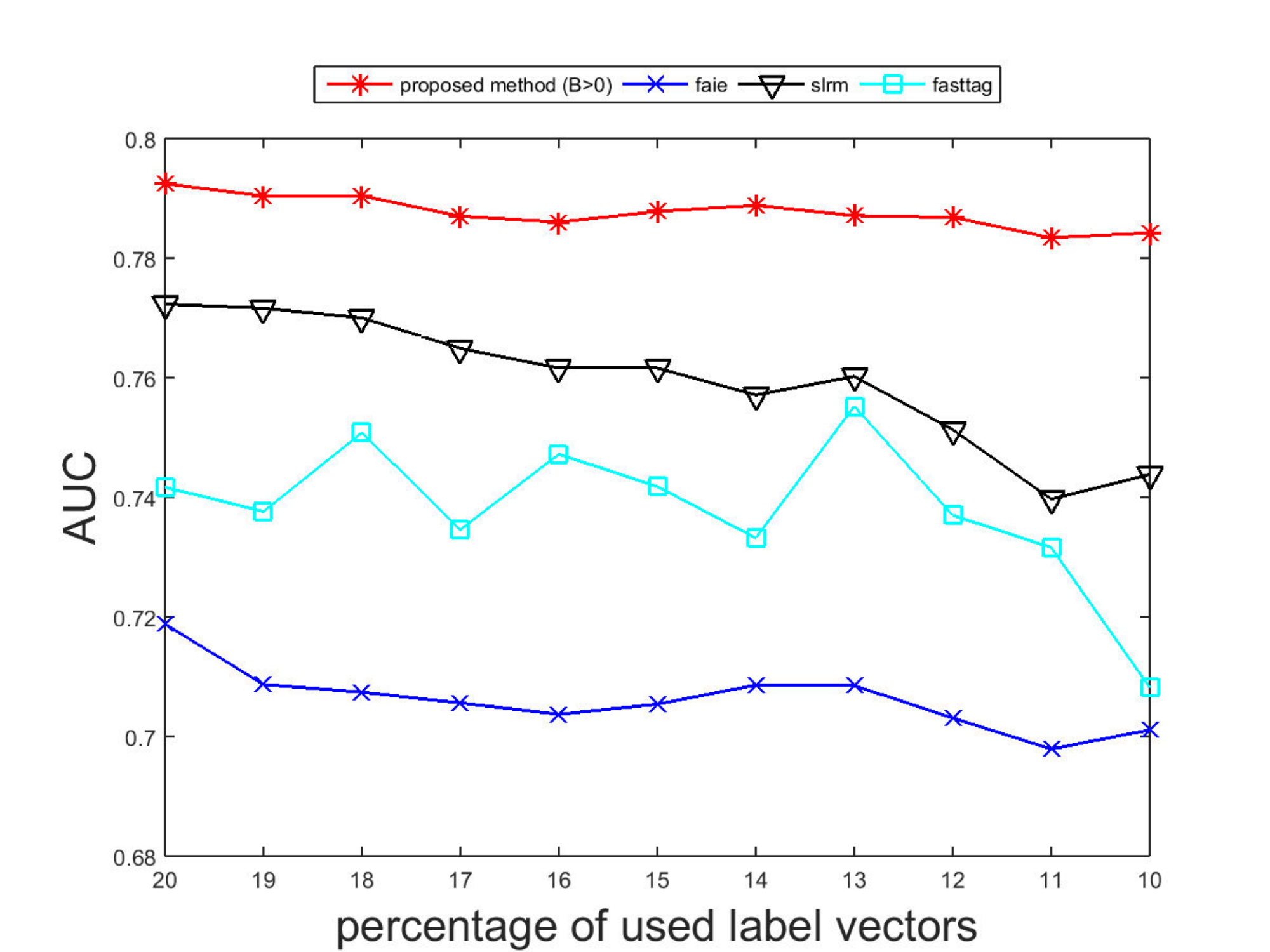}
  \caption*{(a) AUC}
  \label{fig:sfig1}
\end{subfigure}%
\begin{subfigure}{.25\textwidth}
  \centering
  \includegraphics[width=1.05\linewidth]{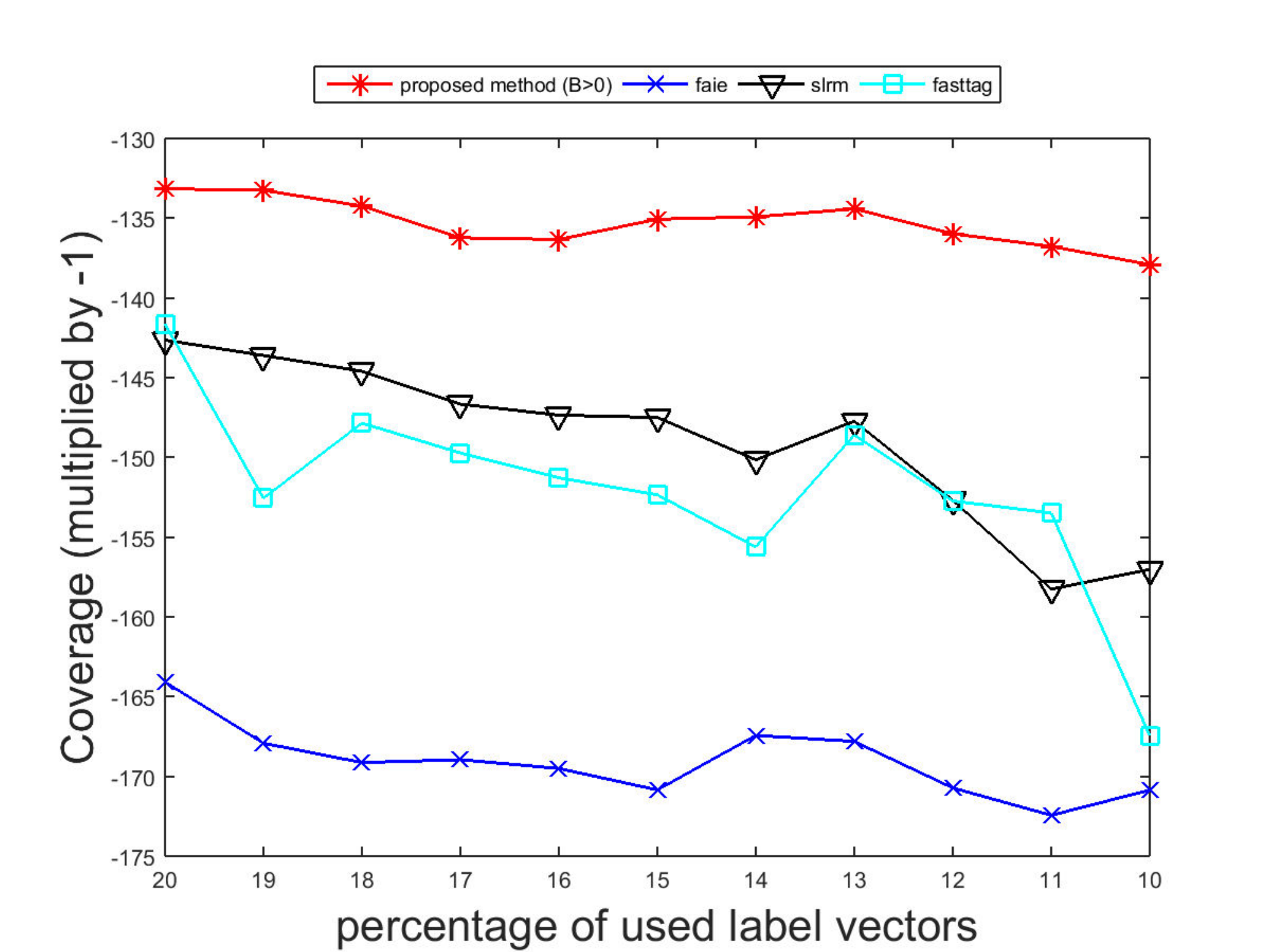}
  \caption*{(b) -Coverage}
  \label{fig:sfig2}
\end{subfigure}
\caption{{Results on espgame, for different percentages of randomly selected label vectors (i.e. simulating the semi-supervised setting).}}
\label{fig:espgame_semisupervised}
\end{figure}

\begin{table}
\caption{Comparison of the results of the proposed ESMC and LEML on one large-scale dataset and two medium-scale datasets.}
\label{table:leml_vs_ESMC}
\begin{tabular}{c|c|c|c}
 & nus-wide & mediamill &delicious\\\hline
LEML \cite{leml} & P@1 $\;\;\;\;\;20.76$ &P@1 $\;\;\;\;\;84.00$&P@1 $\;\;\;\;\;65.66$\\
                          & P@3 $\;\;\;\;\;16.00$ &P@3 $\;\;\;\;\;67.19$&P@3 $\;\;\;\;\;60.54$\\
                          &time (s) $\;\;1097\;\;$ &time (s) $\;N.A.\;$&time (s) $ \;N.A.\;$\\\hline
The proposed ESMC  & P@1 $\;\;\;\;\;\mathbf{40.63}$ &P@1 $\;\;\;\;\;\mathbf{84.82}$&P@1 $\;\;\;\;\;61.54$\\
                    ($B=0$)&P@3 $\;\;\;\;\;\mathbf{28.48}$ &P@3 $\;\;\;\;\;\mathbf{68.80}$&P@3 $\;\;\;\;\;56.12$\\
                          & time (s)$\;$ $\;\mathbf{126}\;\;$ &time (s) $130.65$&time (s)  $\;56.52$\\

\end{tabular}
\end{table}

\begin{table}
\caption{Training time of some methods on different datasets.}
\label{table:run_times}
\begin{tabular}{c||c|c|c|c}
\backslashbox{\tiny{Dataset}}{\tiny{Method}}
&{ESMC (\tiny{$B>0$})}&SLRM \cite{slrm}&FaIE \cite{faie}
&FastTag\cite{fasttag}\\\hline\hline
CAL500  &$\mathbf{1\;s}$&$11\;s$&$\mathbf{1\;s}$&$11\;s$\\
corel5k   &$\mathbf{17\;s}$&$35\;s$&$237\;s$&$120\;s$\\
iaprtc12 &$\mathbf{89\;s}$&$499\;s$&$8204\;s$&$335\;s$\\
espgame &$\mathbf{116\;s}$&$509\;s$&$12910\;s$&$402\;s$\\
delicious &$\mathbf{119\;s}$&$1558\;s$&$16172\;s$&$5094\;s$\\
mediamill &$840\;s$&$1540\;s$&$16255\;s$&$\mathbf{173\;s}$\\
\end{tabular}
\end{table}
The performance of our ESMC method and that of some previous methods are evaluated in different settings. Figs. \ref{fig:cal500_removy}, \ref{fig:corel5k_removy}, \ref{fig:iaprtc12_removy} and \ref{fig:espgame_removy}
provide
the AUC and the coverage as performance measures. Note that we negated the coverage value, so that in all the figures, a larger reported value for an evaluation metric indicates better prediction performance. 
We randomly partitioned instances into training data and test data as mentioned
above. Moreover, to evaluate the ability of different methods
to handle missing labels, a fraction (i.e, 10\%, 20\%, 30\%, 40\%,
and 50\%) of label assignments are randomly removed in the
training set. The horizontal axis in these figures shows the percentages of the removed labels in the training set. Therefore,
the performance of these methods for 0\% removal rate presents
their multi-label classification performance on the corresponding datasets. These figures illustrate that in almost all of the experiments, the proposed ESMC method outperforms the other mentioned methods by a large margin. Moreover, these figures demonstrate that introducing the auxiliary random variables (i.e. experts) improves the prediction performance.
As previously explained, the proposed ESMC and FastTag \cite{fasttag} have explicit mechanisms to handle missing labels.
Moreover, SLRM \cite{slrm} handles missing labels by filling such missing entries with label correlations and intrinsic structure among data \cite{slrm}. However, FaIE \cite{faie} has no explicit mechanism to handle missing labels. Figs. \ref{fig:cal500_removy}, \ref{fig:corel5k_removy}, \ref{fig:iaprtc12_removy}, and \ref{fig:espgame_removy} confirm this notion. These figures demonstrate that the proposed ESMC, SLRM \cite{slrm}, and FastTag \cite{fasttag} can compensate missing labels. However, in the presence of missing labels, the performance of FaIE \cite{faie} is prone to decrease. 

We randomly partitioned instances into training data and test data as mentioned
above. Moreover, to evaluate the ability of different methods
to exploit the unlabeled instances, a fraction (i.e, 10\%-20\%) of label vectors are randomly selected and used in the
training phase. Note that in this setting, we randomly selected some training data and use the label vectors of only these instances (other instances can be considered as unlabeled data), while in previous setting, we randomly selected a fraction of the one entries in the matrix $\mathbf{Y}$ and set this entries to zeros. 
The horizontal axis in Figs. \ref{fig:corel5k_semisupervised}, \ref{fig:iaprtc12_semisupervised}, and \ref{fig:espgame_semisupervised} shows the percentages of the used label vectors in the training set. 
Recall that among these methods, SLRM \cite{slrm} and the proposed ESMC can exploit unlabeled data. These figures illustrate that in almost all of the experiments, the proposed ESMC outperforms the other mentioned methods. 

To evaluate the ability of the proposed ESMC to handle large-scale datasets, we used one large-scale dataset, NUS-WIDE, as well as two medium-scale datasets, mediamill and delicious. 
Table \ref{table:leml_vs_ESMC} compares the performance of our proposed method with that of LEML \cite{leml} (in terms of Precision@1, Precision@3, and training time). In this setting, we ran each algorithm once. In fact, we used the standard training and test data provided for these datasets.
For LEML \cite{leml}, we used the reported results in the article of LEML \cite{leml}, as well as the online-available results \cite{xml_repository}. According to Table \ref{table:leml_vs_ESMC}, for NUS-WIDE, the proposed ESMC is almost ten times faster than LEML \cite{leml}.

Adding the experts make the proposed method capable of filling the missing entries in the label matrix. However, doing so can increase the number of incorrect predicted label assignments. Consequently, adding the experts can dramatically affect the Precision@1 and Precision@3. Thus, in the experiments of Table \ref{table:leml_vs_ESMC}, we used the proposed method with no experts.

\section{Conclusion}
In this paper, we proposed an embedding-based multi-label classifier that models the transformation mapping the feature space to the latent space, as well as the transformation mapping the latent space to the label space, by two sets of stochastic transformations. In this regard, our method is more flexible than state-of-the-art linear approaches and in terms of prediction performance, outperforms them by a large margin. Furthermore, by modeling these mappings using stochastic transformations, our method addresses the problem of neglecting the tail labels, which is not addressed by many state-of-the-art embedding-based multi-label classifiers. 
One of the most important contributions of this paper is to exploit the idea of parameterizing sparse Gaussian processes by pseudo-instances that dramatically decreases the training time of the proposed method. Moreover, our ESMC method uses effective mechanisms to compensate missing labels, and to exploit unlabeled instances.




%


\bibliography{IEEEabrv,main}

\end{document}